\documentclass[twocolappendix,numberedappendix]{emulateapj}
\usepackage{natbib}
\usepackage{graphicx}

\usepackage{mathrsfs}
\usepackage{amsfonts}
\usepackage{amsmath,amssymb}
\usepackage{color}
\usepackage{algorithm}
\usepackage{algorithmic}
\usepackage{bm}
\newcommand{\bL}{\boldsymbol{L}}

\newcommand{\x}{\bm{x}}
\newcommand{\y}{\bm{y}}
\newcommand{\w}{\bm{w}}
\newcommand{\K}{\bm{K}}

\newcommand{\D}{\mathcal{D}}
\newcommand{\T}{\mathrm{T}}

\newcommand{\fK}{\mathbf{K}}
\newcommand{\fO}{\mathcal{O}}

\def\bbbr{{\rm I\!R}} % for expectation
\begin{document}
\title{Nonparametric Bayesian Estimation of Periodic Lightcurves}

\author{Yuyang Wang\altaffilmark{1},
Roni Khardon\altaffilmark{1},
Pavlos Protopapas\altaffilmark{2,3}}
\affil{\altaffilmark{1}Department of Computer Science,
Tufts University, Medford, MA, USA}
\affil{\altaffilmark{2}Harvard-Smithsonian Center for Astrophysics, Cambridge, MA, USA}
\affil{\altaffilmark{3}Institute for Applied Computational Science, Harvard University, Cambridge, MA, USA}
%% \numberofauthors{3}
%\author{
%\alignauthor Yuyang Wang \\
%       \affaddr{Dept. of Computer Science}\\
%       \affaddr{Tufts University}\\
%       \affaddr{Medford, MA 02155, US}\\
%       \email{ywang02@cs.tufts.edu}
%\alignauthor Roni Khardon\\
%       \affaddr{Dept. of Computer Science}\\
%       \affaddr{Tufts University}\\
%       \affaddr{Medford, MA 02155, US}\\
%       \email{roni@cs.tufts.edu}
%\alignauthor Pavlos Protopapas\\
%       \affaddr{Harvard-Smithsonian Center for Astrophysics}\\
%       \affaddr{Harvard University}\\
%       \affaddr{Cambridge, MA 02140, US}\\
%       \email{pprotopapas@cfa.harvard.edu}
%}
% \date{\today}
\begin{abstract}
Many astronomical phenomena  exhibit patterns that have periodic
behavior. An important step when
analyzing data from such processes is the problem of identifying the
period: estimating the period of a periodic function based on noisy
observations made at irregularly spaced time points.  This problem
is still a difficult challenge despite extensive study in different
disciplines.  This paper makes several contributions toward solving
this problem.  First, we present a nonparametric Bayesian model for
period finding, based on Gaussian Processes (GP), that does not make
assumptions on the shape of the periodic function.  As our
experiments demonstrate, the new model leads to significantly better
results in period estimation especially when the lightcurve does not exhibit
sinusoidal shape.  Second, we develop a new algorithm for parameter
optimization for GP which is useful when the likelihood function is
very sensitive to the parameters with numerous
local minima, as in the case of period estimation.  The algorithm
combines gradient optimization with grid search and incorporates
several mechanisms to overcome the high computational complexity of GP.
Third, we develop a novel approach for using domain knowledge,
in the form of a probabilistic generative model, and incorporate it
into the period estimation algorithm.
Experimental results validate our approach showing significant
improvement over existing methods.
\end{abstract}

%[PP: STILL UNRESOLVED
% 1) FONTS IN PLOTS ARE TOO SMALL.
% 2) PROVIDE LIGHTCURVES WHERE OUT METHOD FAILS
% AND LS GETS IT RIGHT ?

% BW:
% 1) I think the fonts are fine (larger is less beautiful). If it is OK, we can add the legend in the caption.
% 2) As far as MACHO is concerned, comparing to LS where LS got right, our method gives a mess. But not sure in OGLE.

%An extensive experimental evaluation iedntifies a robust and efficient
%algorithm for parameter optimization in this context.

%In this paper, we develop novel algorithms
%for period finding using a nonparametric Bayesian model, based on
%Gaussian Processes, that does not make strong assumptions on the
%shape of the periodic function.
%Our paper is the first to
%address this problem from a data mining perspective, adding
%algorithmic and computational considerations and empirical
%comparisons to evaluate ideas and methods. In addition to
%identifying the period, our model and algorithms provide a
%regression estimate of the periodic function.  Experiments using
%both synthetic data and time series data from astrophysics
%demonstrate the performance of the proposed method.

%%introduction
\section{Introduction}
\begin{figure}[h!]
{
\includegraphics[width=0.5\textwidth]{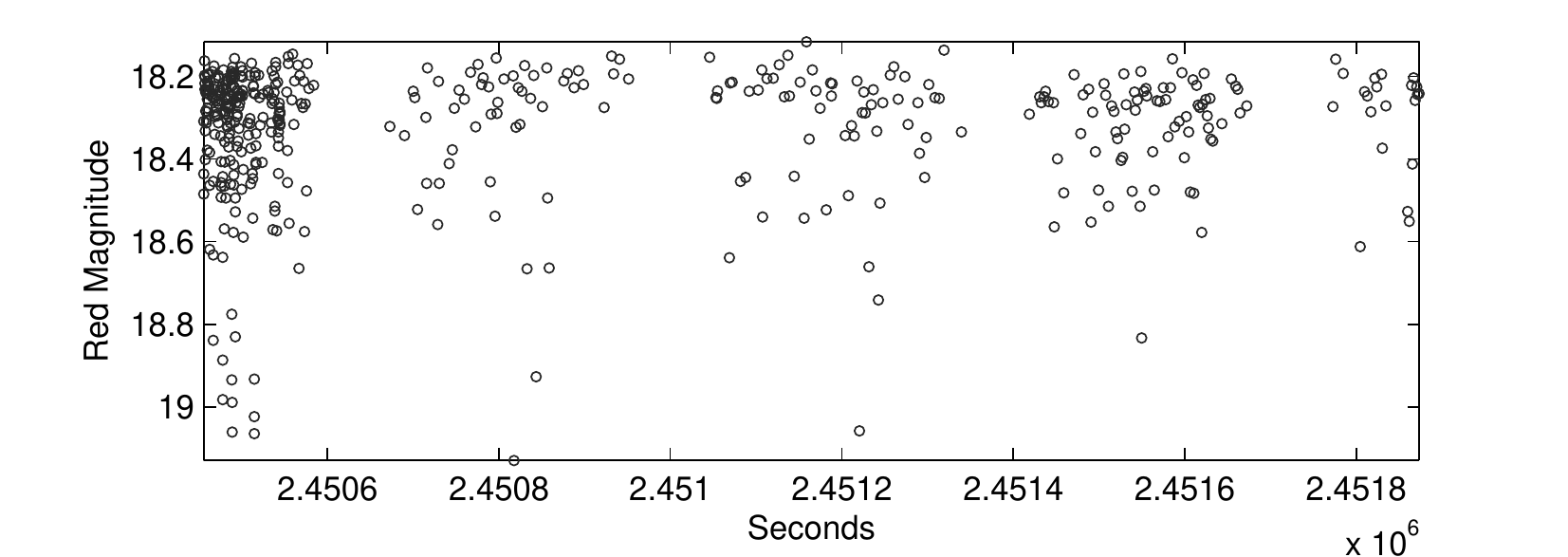}
\includegraphics[width=0.5\textwidth]{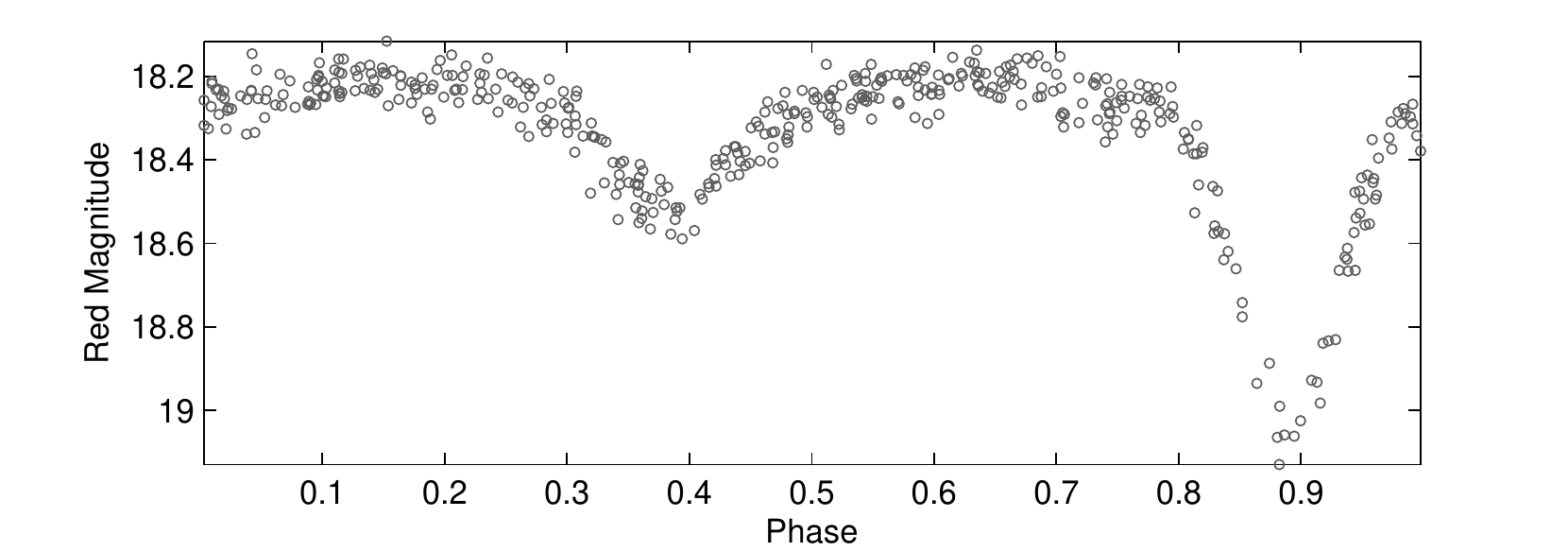}
\caption{Top: brightness of an eclipsing binary star over time;
Bottom: brightness versus phase. } \label{fig:ebdemo}}
\end{figure}

Many astronomical phenomena exhibit periodic behavior.
 Discovering their
period and the periodic pattern they exhibit is an important task
toward understanding their behavior. A significant effort has been devoted to the
analysis of  lightcurves from periodic variable stars. % .
%A star is classified as variable if its apparent brightness as seen from
%Earth changes over time, whether the changes are due to variations
%in the star's actual luminosity, or to variations in the amount of
%the star's light that is blocked from reaching Earth.
%The graph of star brightness, as a function of phase during the cycle, can give
%insight into the nature of these mechanisms within the star, or
%within the star system.
For example, the top part of Figure~(\ref{fig:ebdemo}) shows the magnitude of a light source over
time. The periodicity of the light source is not obvious before we
fold it. However, as the bottom part illustrates, once folded with
the right period we get convincing evidence of periodicity. The
 object in this figure is classified as an eclipsing binary
(EB). Other sources show periodic variability due to processes internal to the
star \citep{Petit1987}.

The problem of period estimation from noisy and irregularly sampled
observations has been studied before in several disciplines. Most
approaches identify the period by some form of grid search. That is,
the problem is solved by
evaluating a criterion $\Phi$ at a set of trial periods $\{p\}$
%comparing the observations folded at each
%trial periods to a criterion, $\Phi(p)$,
and
selecting the period $p$ that yields the best  value for $\Phi(p)$.
The commonly-used techniques vary in the form and parametrization of
$\Phi$, the evaluation of the fit quality between model and data,
the set of trial periods searched, and the complexity of the resulting
procedures.
Two methods we use as baselines in our study are the LS
periodogram \citep{scargle1982studies, reimann1994frequency} and the
phase dispersion minimization (PDM) \citep{stellingwerf1978period},
both known for their success in empirical studies. The LS method is
relatively fast and is equivalent to maximum likelihood estimation
under the assumption that the function has a sinusoidal shape. It
therefore makes a strong assumption on the shape of the underlying
function. On the other hand, PDM makes no such assumptions and is
more generally applicable, but it is slower and is less often used
in practice. A more extensive discussion of related work is given in
Section~\ref{sec:relatedwork}.

The paper makes several contributions toward solving the period
estimation problem.  First, we present a new model for period finding,
based on Gaussian Processes (GP), that does not make strong
assumptions on the shape of the periodic function.
In this context,
the period is a hyperparameter of the covariance function of the GP and
accordingly the period estimation is cast as a model selection problem for
the
%corresponding
GP.
%[PP. I DO NOT THINK WE NEED TO SAY THE FOLLOWING]
%To our knowledge this is the first use of GP, which provide a well founded
%and flexible framework, for period estimation.
%[REFERENCE A STATISTICS BOOK OR A REVIEW ARTICLE
% FOR GENERATIVE MODEL, LATENT VARIABLES, MARGINAL
% LIKEHOOD
As our experiments demonstrate, the new model leads to significantly better results
compared to LS when the target function is non-sinusoidal. The model
also significantly outperforms PDM when the sample size is small.

Second, we develop a new algorithm for period estimation within the GP
model. In the case of period estimation the likelihood function is
not a smooth function of the period parameter. This results in a difficult
estimation problem which is not well explored in the GP literature
\citep{rasmussen2005gaussian}.  Our algorithm combines gradient
optimization with grid search and incorporates several mechanisms to
improve the complexity over the naive approach.

 In particular we propose and evaluate:  an approximation using a two level
grid search, approximation using limited cyclic optimization, a method
using sub-sampling and averaging, and a method using low-rank Cholesky
approximations.
 An extensive experimental evaluation using artificial
data identifies
the most useful approximations and yields a robust algorithm for
period finding.
%Our estimation algorithm is developed in the context of
%period finding but the ideas are general and can be used in other
%applications of GP.

Third, we develop a novel approach for using astrophysics knowledge, in the
form of a probabilistic generative model,
and incorporate it into the
period estimation algorithm.
In particular, we propose to employ the generative model
to bias the selection of periods by using it as a prior over periods or
as a post-processing
selection criterion choosing among
periods ranked highly by the GP.
%Generative models have been previously
%developed in  astrophysical applications using existing datasets
%and are readily applicable for period finding.
The resulting algorithm is applied and evaluated on astrophysics
data showing significantly improved performance over previous work.

%The remainder of the paper is organized as follows. Section 2
%provides some preliminaries on Gaussian processes, formulates the
%problem and describes the basics of model selection for GP. The
%algorithms are presented in Section 3 and the experimental results
%are reported in Section 4. Related work is discussed in Section 5
%and the final section concludes with a discussion and outlines ideas
%for future work.

% if including should cite as in
%Section~\ref{sec:prelim}

The next section provides some technical background and defines the period
estimation problem as GP inference. The following three sections present our
algorithm, report on experiments evaluating it and applying it to
astrophysics data, and discuss related work. The final section concludes with
a summary and directions for future work.

% describe the model and corresponding learning algorithm
\section{Preliminaries: GP for Period Finding}
\label{sec:prelim}

This section provides technical background on GPs and their
optimization procedures and defines the period finding problem in this
context.

Throughout the paper, scalars are denoted using italics, as in
$x,y\in\bbbr$; vectors and matrices use lowercase and capital bold
typeface, as in $\x,\y,\mathbf{K},\mathbf{A}$, and $x_i$ denotes the
$i$th entry of $\x$. For a vector $\x$ and real valued function
$f:\bbbr\rightarrow \bbbr$, we extend the notation for $f$ to
vectors so that $f(\x) = [f(x_1),\cdots, f(x_n)]^\T$ where the
superscript $\T$ stands for transposition. $\mathbb{I}$ is the
identity matrix.

\subsection{Gaussian Processes}
%[PP: I HAVE TO SAY BERNIES LAST VERSION
% WAS EASIER TO FOLLOW THAN THIS
%
This section gives a brief review of Gaussian processes regression.
A more extensive introduction can be found in
\citep{rasmussen2005gaussian,bishop2006pattern}.

We start with the following regression model,
\begin{equation}
\label{eq:basic}
  y = f_{\w}(\x) + \epsilon
\end{equation}
where $f_{\w}(x)$ is the regression function with parameter $\w$ and $\epsilon$ is iid Gaussian noise. For example, in linear regression
$f_{\w}(\x) = \w^T\x$
and therefore $y \sim N(\w^T\x^*,1/\sigma^2) $.
Given the data $\D = \{\x_i, y_i\}, i = 1, \cdots, N$, one wishes to infer $\w$ and the basic approach is to maximize the likelihood $\mathcal{L}(\w, \D) = \Pr(\D|\w)$.

In Bayesian statistics, the parameter $\w$ is assumed to have a prior
probability $\Pr(\w)$ which encodes the prior belief on the parameter. The inference task becomes calculating the posterior distribution over $\w$, which, using the Bayesian formula,
is given as
\begin{equation} \Pr(\w|\D) \propto \Pr(\D|\w)\Pr(\w). \end{equation}
The predictive distribution for a new observation $\x^*$ is given by
\begin{equation}
\Pr(f(\x^*)|\D) = \int\Pr(f(\x^*)|\w)\Pr(\w|\D)d\w.\end{equation}
Returning to linear regression, the common model assumes that the prior for $\w$ is a zero-mean multivariate Gaussian distribution, and the posterior turns out to be multivariate Gaussian  as well.
% RK: I looked at adding the formulas here - this will require several more bits of notation to
% define the formulas  which will then not be used in the paper. I think this will harm more than help
% but if you feel strongly about it I can add the formulas.
%[PP: I STILL THINK THAT IF WE REFER TO REGRESSION
% IT WOULD BE INSTRUCTIVE TO BE AS EXPLICIT AS POSSIBLE
%
%and $Pr(f(\x^*)|\w) = N(\w^T\x^*,1/sigma^2) $.
In contrast with many Bayesian formulations, the use of GP often allows for simple inference or calculation of desired quantities because of properties of multivariate Gaussian distributions and corresponding facts from linear algebra.

This approach can be made more general using a
nonparametric Bayesian model. In this case we replace the parametric
latent function $f_{\w}$ by a stochastic process $f$ where $f$'s prior is given by a Gaussian process. A GP is specified by a mean function (assumed to be zero in this paper) and covariance function
$\mathcal{K}(\cdot,\cdot)$. This allows us to specify a prior over functions $f$ such that the distribution induced by $f$ over any finite sample is normally distributed.
%More precisely, a Gaussian process is a stochastic process whose finite distribution is a multivariate Gaussian distribution. We can think of this as extending the parametric model $f_{\w}$ with a prior multivariate Gaussian distribution on $\w$ to the infinite dimension where we place a Gaussian process prior on $f(\x)$ directly.
More precisely, the GP regression model
with zero mean and covariance
function $\mathcal{K}(\cdot,\cdot)$
is as follows.
Given sample points $[\x_1,\ldots,\x_n]^T$
%the data set $\mathcal{D}=\{\x_i,y_i\},i=1,\cdots,N$
let
$\fK=(\mathcal{K}(\x_i,\x_j))_{i,j}$.
The induced distribution on the values of the function at the sampling points is
\begin{equation}
\mathbf{f} \triangleq
[f(\x_{_1}),\cdots, f(\x_{_N})]^\T\sim\mathcal{N}(\mathbf{0},\fK),
\end{equation}
where $\mathcal{N}$ denotes the multivariate normal distribution.
Now assuming that $y_i$ is generated from $f(\x_{_i})$ using iid noise as in Equation~(\ref{eq:basic})
and denoting
$\y = [y_1,\ldots,y_n]^T$ we get that $\y \sim\mathcal{N}(\mathbf{0},\fK+ \sigma^2\mathbb{I})$
and the joint distribution is given by
%[PP: BERNIE HAD A NICE REFERENCE HERE AND A NICE EXPLANATION
%
%
%Normally, evaluating the posterior distribution for Bayesian inference is hard or sometimes yields no analytic solution. But GP provides a framework of easy inference thanks to results in linear algebra. To see this, we can write the join probability as
\begin{equation}
\begin{bmatrix}
  \mathbf{f} \\
  \y
\end{bmatrix} \sim \cal{N}\left(\begin{bmatrix}
  \mathbf{0}\\
  \mathbf{0}\\
\end{bmatrix}, \begin{bmatrix}
  \fK & \fK\\
  \fK & \fK + \sigma^2\mathbb{I}
\end{bmatrix} \right).
\end{equation}
Using properties of multivariate Gaussians
we can see that
the posterior distribution $\mathbf{f}|\y$ is given by

\begin{equation}
  \Pr(\mathbf{f}|\D) = \mathcal{N}(\fK \,(\sigma^2\mathbb{I}+\fK)^{-1} \, \y \, , \, \sigma^2 \,\,(\sigma^2\mathbb{I}+\fK)^{-1}\fK).
\end{equation}

Similarly, the predictive distribution
for some test point $\x_*$ distinct from
the training examples is given by
%\begin{equation}
\begin{equation}
\label{eq:predictive}
  \begin{split}
    &\Pr(f(\x_*)|\x_*,\D)= \int \Pr(f(\x_*)|\x_*, f)\Pr(f|\D)df\\
    &= \mathcal{N}\bigg(\mathbf{k}(\x_*)^\T(\sigma^2\mathbb{I}+\fK)^{-1}\y, \\
    &\qquad\qquad\mathcal{K}(\x_*,\x_*)-\mathbf{k}(\x_*)^\T(\sigma^2\mathbb{I}+\fK)^{-1}\mathbf{k}(\x_*)\bigg)
  \end{split}
%\end{equation}
\end{equation}

where $\mathbf{k}(\x_*) =
[\mathcal{K}(\x_1,\x_*),\cdots,\mathcal{K}(\x_N,\x_*)]^\T$.

\begin{figure}[th]
{
\includegraphics[width=0.5\textwidth]{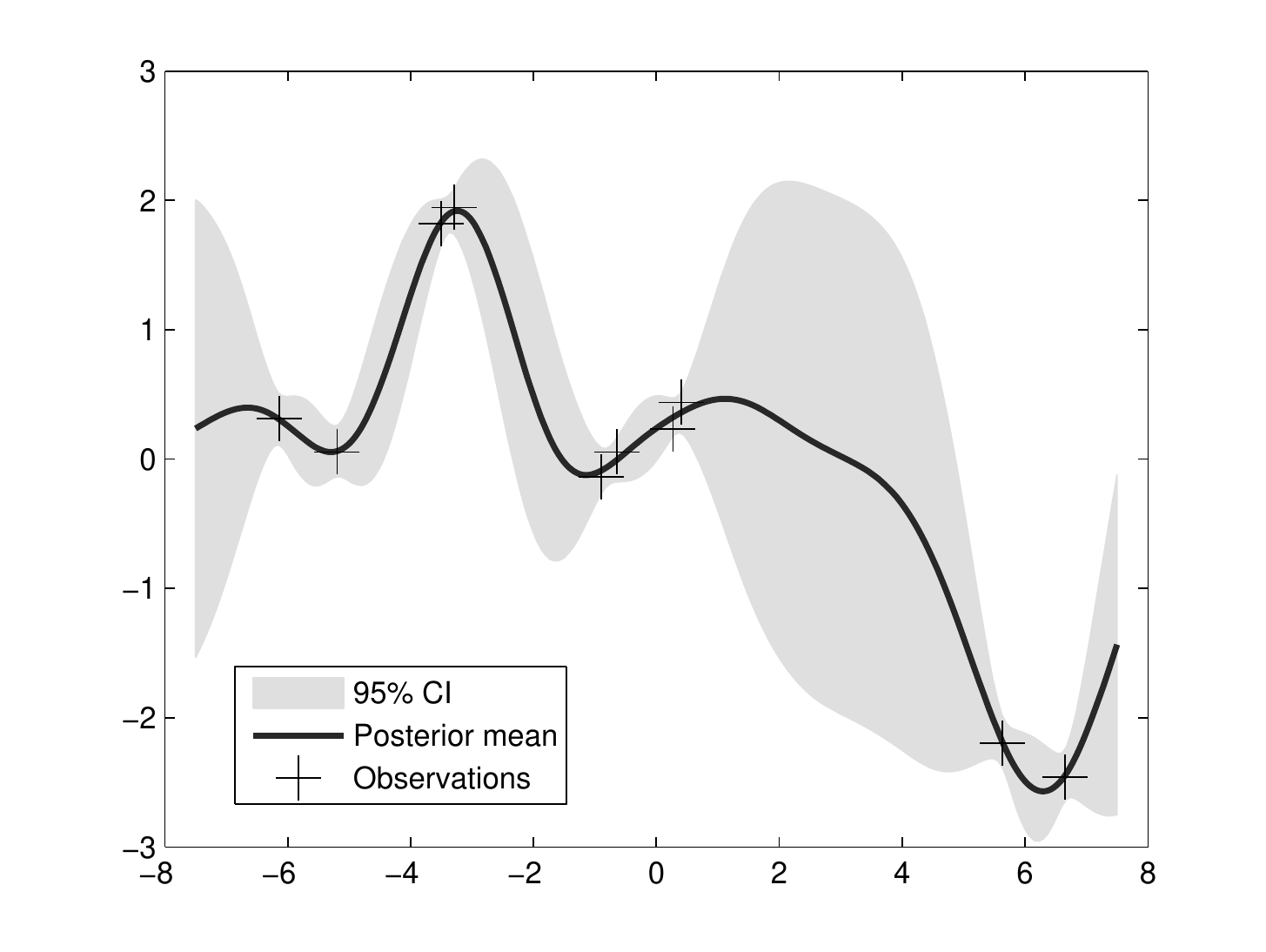}
\caption{Illustration of prediction with GP regression.
The data points $\mathcal{D}=\{\x_i,y_i\}$ are given by the crosses.
The shaded area represents the pointwise 95\% confidence region of the predictive distribution.
As can be seen from Equation~\ref{eq:predictive},
GP regression can be seen to perform a variant of kernel regression where $f(\x_*)$ is a weighted average of all the measurements $\y$. While the values of the weights are obscured because of the inverse of the covariance matrix in that expression, one can view this roughly by an analogy to nearest neighbor regression where the mean of $f(\x*)$ is affected more by the measurements whose sampling points are close to $\x^*$
%due to the smoothness of the function (which is encoded in the prior covariance function). Similarly,
and the variance of $f(\x^*)$ is small if $\x^*$ is surrounded by measurements.
A deeper discussion of the equivalent kernel is given in \citep{rasmussen2005gaussian}.
} \label{fig:gppreddemo}}
\end{figure}

Figure~\ref{fig:gppreddemo} illustrates GP regression, by showing how a finite sample induces a posterior over functions and their values for new sample points.

%Furthermore, we can express a zero mean Gaussian process as a
%distribution on functions $f$ with the following
%probability
%\begin{equation}
%  \label{eqn:fix1}
%  \begin{split}
%    f &\sim \exp\left\{-\frac{1}{2}\|f\|_\mathcal{K}^2\right\}
%  \end{split}
%\end{equation}
%where $\|\cdot\|_\mathcal{K}$ denotes the norm in a Reproducing
%Kernel Hilbert Space (RKHS) with kernel $\mathcal{K}$.

\subsection{Problem Definition}

%Let $f(x)$ be the true value of the phenomenon of interest at time
%$x$. For instance, in astrophysics, $f(x)$ is the measurement of
%``brightness'' taken over time. A graph of $f$, as a function of
%phase, is called the true ``light curve'' of the star in the
%astrophysics literature. Observations $y_i$ are made at respective
%times $x_i$ and the data set is given as $\{x_i,y_i\},
%i\in\{1,2,\cdots, N\}$. We assume that the data is generated
%according to the model

\begin{figure*}[t]
{
\begin{center}
        \begin{minipage}{0.45\textwidth}
        \centering
        \includegraphics[width=\textwidth]{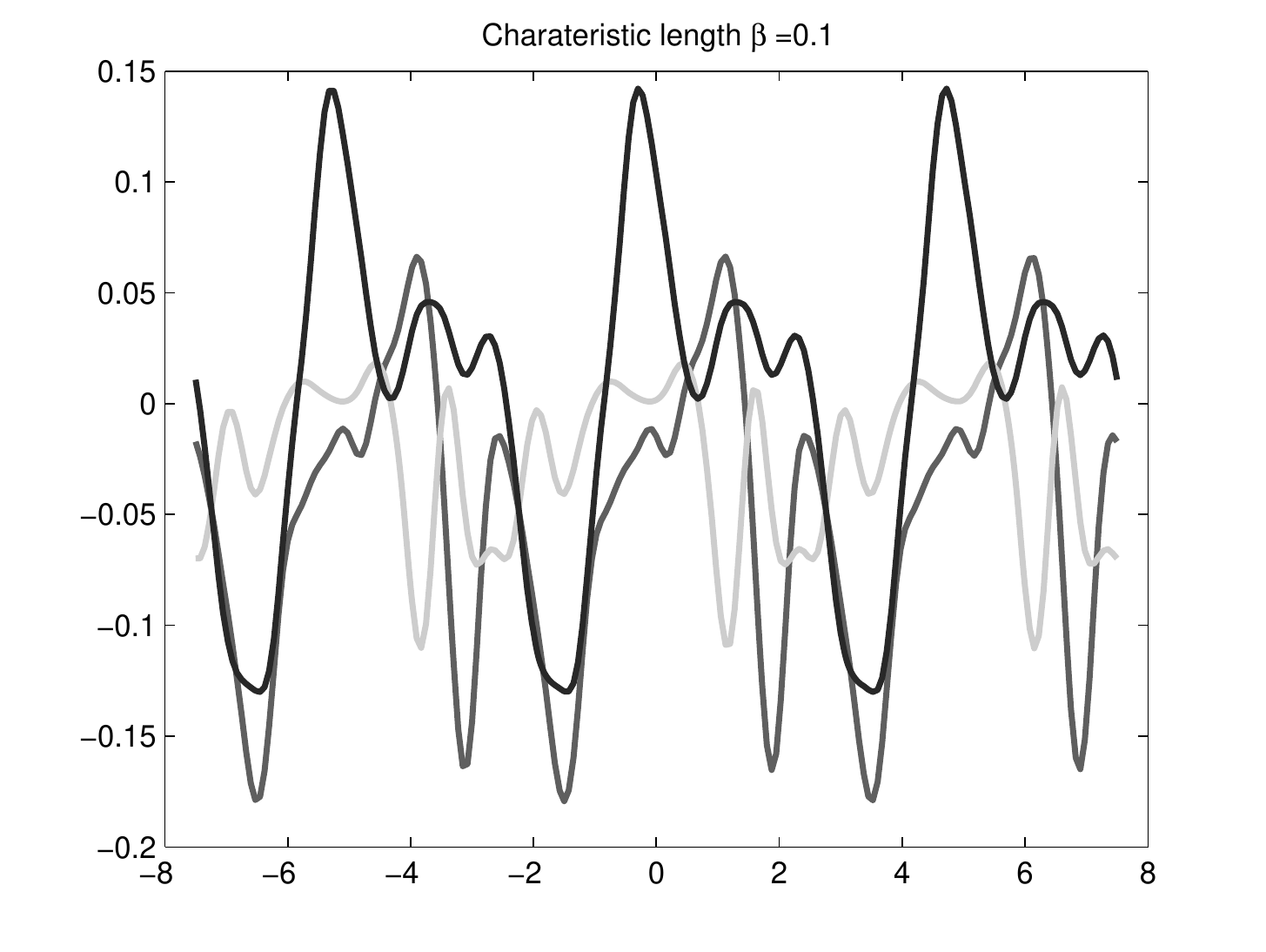}
        \end{minipage}%\hfill
        \begin{minipage}{0.45\textwidth}
        \centering
        \includegraphics[width=\textwidth]{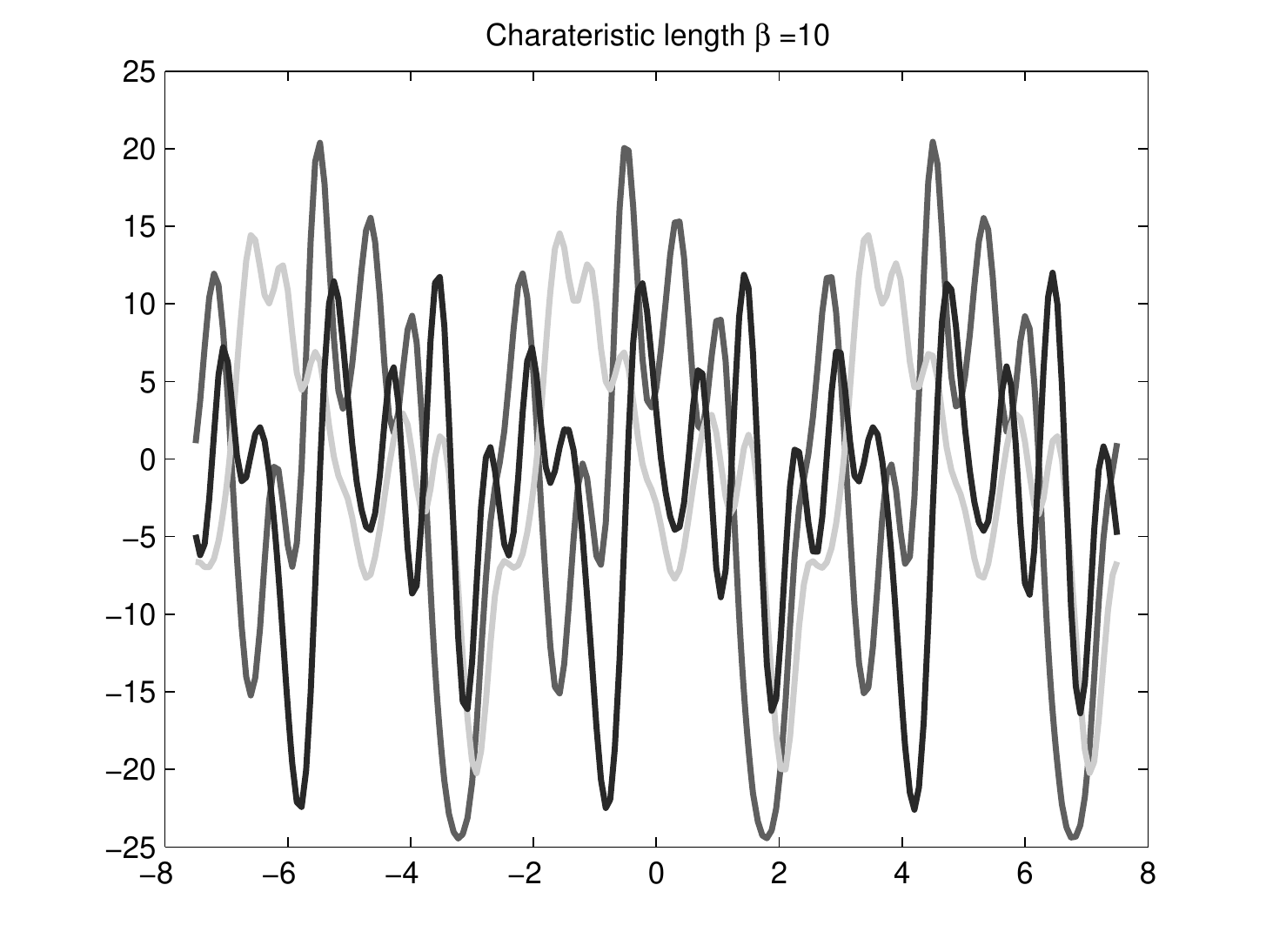}
        \end{minipage}
\\
        \begin{minipage}{0.45\textwidth}
        \centering
        \includegraphics[width=\textwidth]{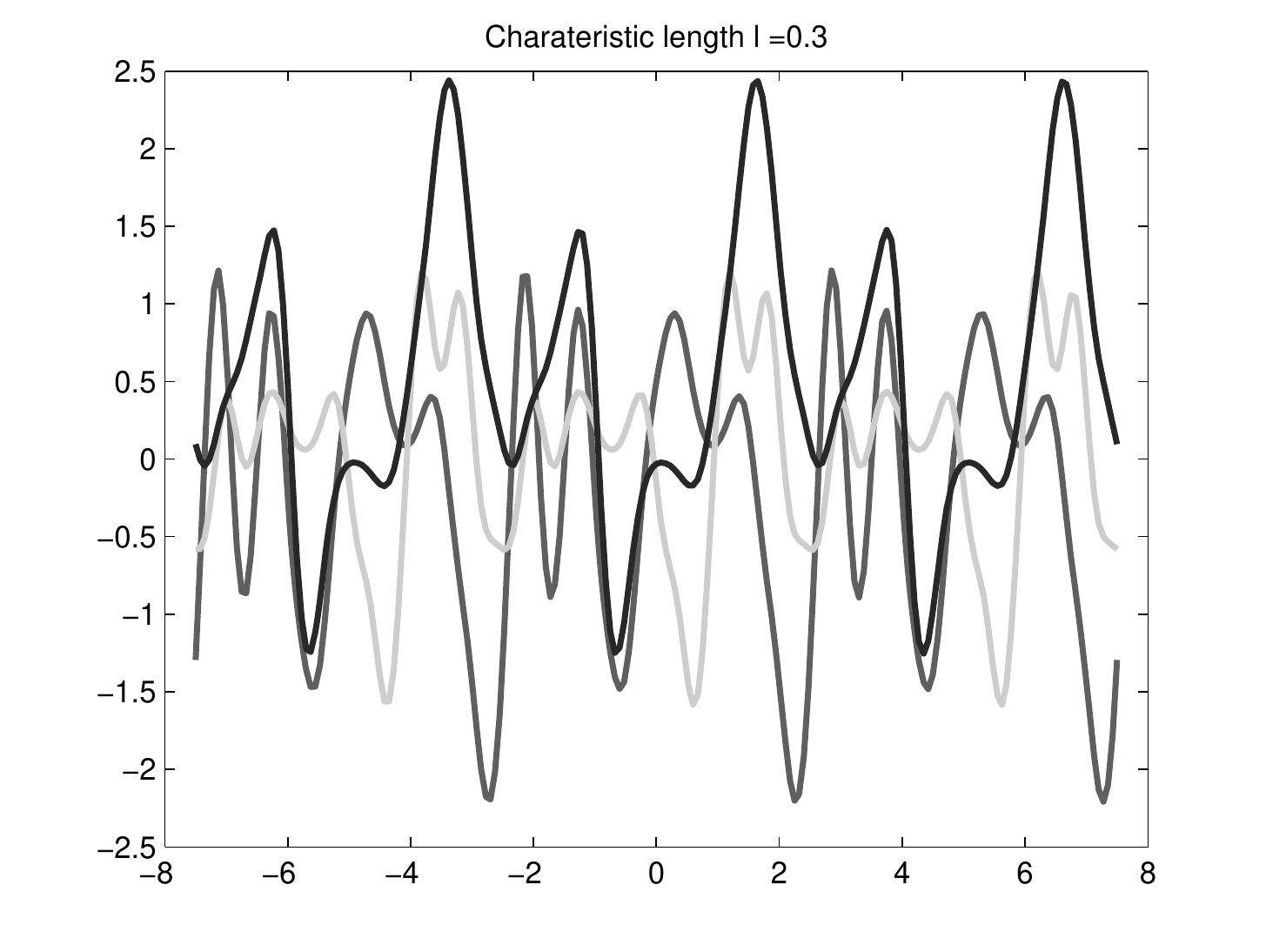}
        \end{minipage}%\hfill
        \begin{minipage}{0.45\textwidth}
        \centering
        \includegraphics[width=\textwidth]{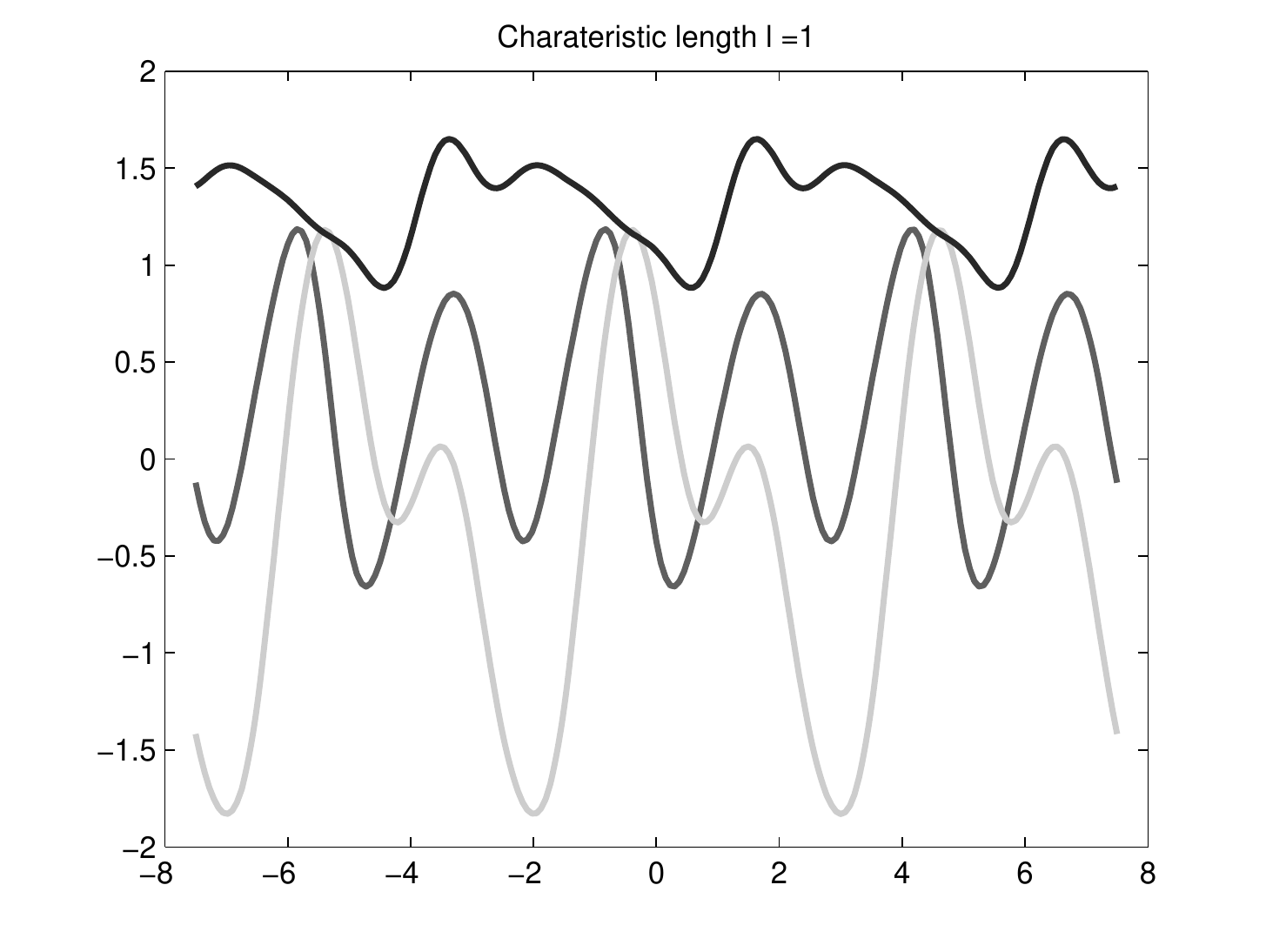}
        \end{minipage}
\end{center}
\caption{Sample functions from a GP with covariance function in
  Equation~(\ref{eqn:pk}) where the period is fixed to be 5, i.e. $w=0.2$. Top row: $\beta = 0.1$ vs $\beta = 10$ while $\ell$ is fixed to be 1. Bottom row:
$\ell = 0.3$ vs $\ell = 1$ with $\beta = 0.3$. }
%\label{fig:gpsine}}
\label{fig:prbf}}
\end{figure*}

In the case of period estimation the sample points
$\x_i$ are scalars $x_i$ representing the corresponding time points,
and we denote
$\x = [x_1,\ldots,x_n]^T$.
The underlying function $f(\cdot)$ is periodic with unknown
period $p$ and corresponding frequency $w=1/p$.
To model the periodic aspect we use a GP with a periodic covariance function

%Thus our mode assumes that
%\begin{equation}
%\label{eqn:npg}
%  y_i = f(x_i) + \epsilon_i
%\end{equation}
%$f$ is generated from a zero mean GP
%with a kernel
\begin{equation}
\label{eqn:pk} \mathcal{K}_{\bm{\theta}}(x_i, x_j) =
\beta\exp{\left\{-\frac{2\sin^2\left(w\pi(x_i -
x_j)\right)}{\ell^2}\right\}},
\end{equation}
where the set of hyperparameters\footnote{
Typically, in a hierarchical model, the parameters of the top level (e.g. parameters of the prior) that affect the next level are called hyperparameters.
In
GP regression, the parameter is the regression function $f$ and the hyperparameters are the the parameters of covariance function.}
 of the covariance function is given by $\bm{\theta} =
\{\beta, w, \ell\}$. It can be easily seen that any $f$ generated by
$\mathcal{K}_{\bm{\theta}}$ is periodic with period $1/w$.
Figure~(\ref{fig:prbf}) illustrates the role of the other two hyperparameters.
%[PP: CHANGE THE BOTTOM TWO PANELS TO BE
% beta=0.1, l=0.3 and l=1
We can see that $\beta$ controls the magnitude of the sampled functions.
At the same time, $\ell$ which is called characteristic length determines how sharp the variation is between two points.

In our problem each star has its own period and shape and therefore
each has its own set of hyperparameters.
Our model, thus, assumes that the following generative process is the one
producing the data.
For each time series $j$ with arbitrary sample points
$\x^j= [x_1^j, \cdots, x_{N_j}^j]^T$, we first draw

\begin{equation}
f_j | \theta_j
\sim
\mathcal{GP}(0, \mathcal{K}_{\bm{\theta}_j}).
\end{equation}
Then, given $\x^j$ and $f_j$ we sample the observations
\begin{equation} \y^j \sim
  \mathcal{N}(f_j(\x^j), \sigma^2\mathbb{I}).
  \end{equation}

\noindent Denote the complete set of parameters by $\mathcal{M} = \{\bm{\theta}, \sigma^2\}$.
For each time series $j$, the inference task is to select the correct model for the data
$\{\x^j,\y^j\}$, that is, to find $\mathcal{M}$ that best describes the
data.  This is the main computational problem studied in this paper.
The next subsection reviews two standard approaches for this
problem.

Before presenting these we clarify two methodological issues.
First,
notice that our model assumes homogeneous noise $\mathcal{N}(0, \sigma^2)$, i.e.  the observation error for each $x_i$ is the same. Experimental results on the astronomy data
(not shown here) show that
$\sigma^2$ estimated from the data is very close to the mean of the recorded observation errors, and therefore there is no advantage in explicitly modeling the recorded observation errors.

Second, as defined above our task is to find the full set of parameters $\mathcal{M}$.
Therefore, our framework and induced algorithms can estimate the underlying function, $f$,
through the posterior mean $\hat{f}$, and thus yield a solution for the
regression problem -- predicting the value of the function at unseen sample points.
However, our main goal and interest in solving the problem
is to infer the frequency $w$ where the other parameters are less important.
Therefore, a large part of the evaluation in the paper focuses on accuracy in identifying the frequency, although we also report results on prediction accuracy for the regression problem.

\subsection{Model selection}
\subsubsection{Marginal Likelihood}
The standard Bayesian approach is to identify the hyper-parameters that
maximize the marginal likelihood.
%, where the ``marginal'' means that
%we integrate out the latent function associated with the data.
More
precisely, we try to find $\mathcal{M}^*$ such that
\begin{equation}
\mathcal{M}^{*} =
\underset{\mathcal{M}}{\text{argmax}}\left[\log\left[ \Pr(\y|\x;
\mathcal{M}) \right]\right]
\end{equation}
where the marginal likelihood is given by
\begin{equation}
\label{eqn:malike}
\begin{split}
\log\Pr(\y|\x; \mathcal{M}) &= \log\left(\int \Pr(\y|f, \x; \mathcal{M})\Pr(f|\x; \mathcal{M})\, df\right)\\
&= -\frac{1}{2}\y^T(\K + \sigma^{2}\mathbb{I})^{-1}\y\\
&\quad -\frac{1}{2}\log|\K+ \sigma^{2}\mathbb{I}|^{-1} -
\frac{n}{2}\log 2\pi
\end{split}
\end{equation}
and Equation~(\ref{eqn:malike}) holds because $\y\sim \mathcal{N}(\mathbf{0}, \K +
\sigma^{2}\mathbb{I})$~\citep{rasmussen2005gaussian}.  Typically, one can
optimize the marginal likelihood by calculating the partial derivative of the
marginal likelihood w.r.t.\ the hyper-parameters and optimizing the
hyper-parameters using gradient based search \citep{rasmussen2005gaussian}. As
we show below, gradients alone cannot be used to solve our problem completely
and therefore our algorithm elaborates and improves over this approach.
We do, however, use the conjugate gradients optimization as a basic step in our
algorithm.
The partial derivative of
Equation~(\ref{eqn:malike}) w.r.t.\ the parameter $\theta_j$ is
~\citep{rasmussen2005gaussian}
\begin{equation}
\begin{split}
  \frac{\partial}{\partial\theta_j}\log \Pr(\y|\x;\mathcal{M}) &= \text{Tr}\left(\left(\alpha\alpha^T - \K_\sigma^{-1}\right)\frac{\partial{\K_\sigma}}{\partial\theta_j}\right)
\end{split}
\end{equation}
where $\K_\sigma = \K+ \sigma^{2}\mathbb{I}$ and $\alpha =
\K_\sigma^{-1}\y$.

\subsubsection{Cross-Validation}
An alternative approach~\citep{rasmussen2005gaussian} picks
hyperparameter $\mathcal{M}$ by minimizing the empirical loss on a
hold out set. This is typically done with a leave-one-out (LOO)
formulation, which uses a single observation from the original sample as the validation data, and the remaining observations as the training data. The process is repeated such that each observation in the sample is used once as the validation data. To be precise, we choose the hyperparameter $\cal{M}^*$ such that

\begin{equation}
\label{eqn:cv} \mathcal{M}^{*} =
\underset{\mathcal{M}}{\text{argmin}}\sum_{i=1}^{n}(y_{i} -
\hat{f}_{-i}(x_{i}))^{2}
\end{equation}
where $\hat{f}_{-i}$ is defined as the posterior mean given the data
$\{\x_{-i}, \y_{-i}\}$ in which the subscript $-i$ means all but the
$i$th sample, that is,
\begin{equation}
\hat{f}_{-i}(x) = \mathcal{K}(\x_{-i}, x)^{T}\left(\K_{-i} +
\sigma^2\mathbb{I}\right)^{-1}\y_{-i}.
\end{equation}
It can be shown that
this computation can be simplified \citep{rasmussen2005gaussian}
using the fact that
\begin{equation}
  y_{i} - \hat{f}_{-i}(x_{i}) = \frac{\left[(\K + \sigma^2\mathbb{I})^{-1}\y\right]_i}{\left[(\K + \sigma^2\mathbb{I})^{-1}\right]_{ii}}
\end{equation}
where $[\cdot]_i$ is the $i$th entry of the vector and
$[\cdot]_{ii}$ denotes the $(i,i)$th entry of the matrix.

\section{Algorithm}

\begin{figure*}[t]
{
        \begin{minipage}{0.48\textwidth}
        \centering
        \includegraphics[height=6cm]{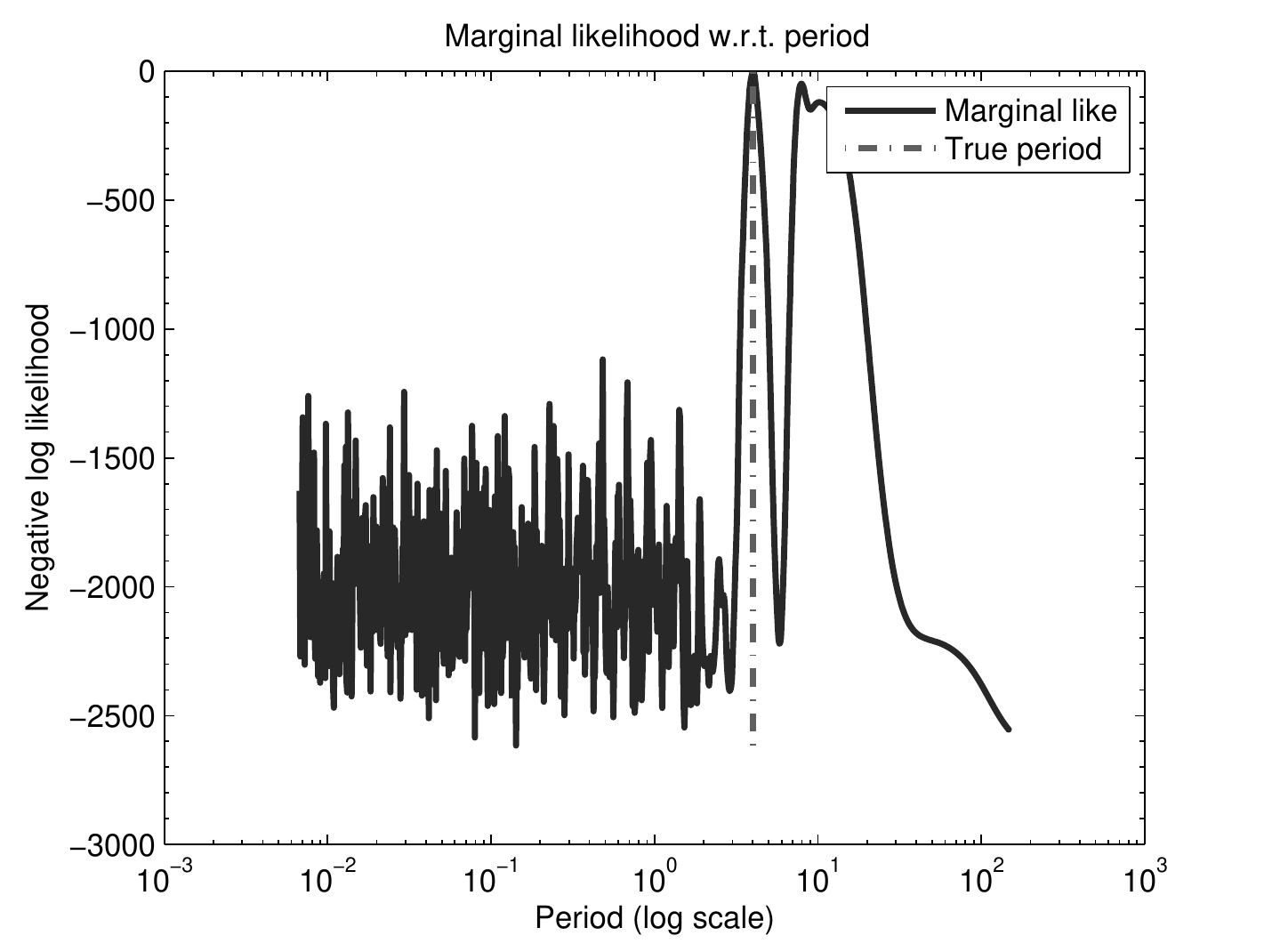}
        \end{minipage}%\hfill
        \begin{minipage}{0.48\textwidth}
        \centering
        \includegraphics[height=6cm]{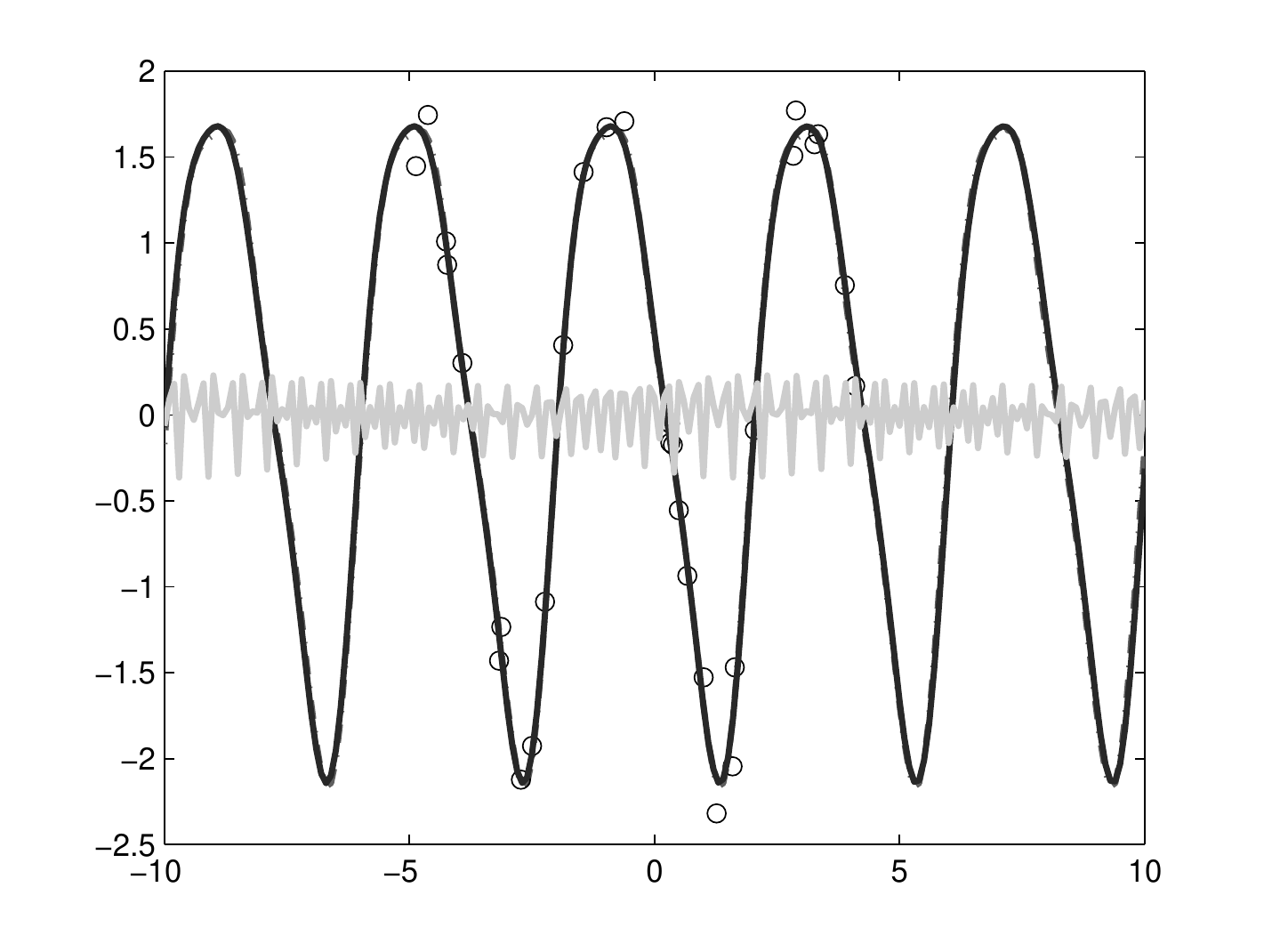}
        \end{minipage}
\caption{Illustration of sensitivity of the marginal likelihood. A light curve is generated using the GP model with parameters $\beta=1$, $w=0.25$, and $\ell=1$.
Left: The marginal likelihood function versus the period,
where the dotted line indicates the true period.
Right: The black circles are the observations and the dotted line (covered by the dark estimated curve)
is the true function. The dark line which covers the true curve and the
light line are the learned regression functions
given two different starting points of $w$.}
\label{fig:1ab}}
\end{figure*}

\begin{figure*}[t]
\centering
\fbox{
\begin{minipage}{0.95 \linewidth}
%\vspace{0.3cm}
\algsetup{indent=2em} %\newcommand{\factorial}{\ensuremath{\mbox{\sc Algorithm summary}}}
%\begin{algorithm}[h!] \caption{\sc Frequency estimation algorithm using GP regression model}\label{alg:gmt}
%\begin{algorithm}[h!]
%\caption{\sc Frequency estimation using GP}\label{alg:gmt}
%\begin{algorithm}[h!]
%\caption{\sc Frequency estimation using GP}\label{alg:gmt}
\begin{algorithmic}[1]
\STATE Initialize the parameters randomly.
\REPEAT
\STATE Jointly find $\tilde{w}, \beta^*, \ell^*, \sigma^*$ that maximize
Equation~(\ref{eqn:malike}) using conjugate gradients.
\FORALL{$w$ in a coarse grid set $\mathcal{C}$}
\STATE Calculate the marginal likelihood Equation~(\ref{eqn:malike}) or the LOO
Error Equation~(\ref{eqn:cv}) using $\beta^*, \ell^*, \sigma^*$.
\ENDFOR
\STATE Set $w$ to the best value found in the \textbf{for} loop.
\UNTIL{Number of iterations reaches $L_1$ ($L_1=2$ by default)}
\STATE Record the Top $K$ ($K=10$ by default) frequencies $\mathcal{W}^*$
found in the last run of \textbf{for} loop (lines 4-6).
\REPEAT
\STATE Jointly find $\tilde{w}, \beta^*, \ell^*,
\sigma^*$ that maximize Equation~(\ref{eqn:malike}) using conjugate
gradients.
\FORALL{$w$ in a fine grid set $\mathcal{F}$ that covers
$\mathcal{W}^*$}
\STATE Calculate the marginal likelihood Equation~(\ref{eqn:malike})
or the LOO Error Equation~(\ref{eqn:cv}) using
$\beta^*, \ell^*, \sigma^*$.
\ENDFOR
\STATE Set $w$ to the best value found in the \textbf{for} loop.
\UNTIL{Number of iterations
reaches $L_2$ ($L_2=2$ by default)}
\STATE Output the frequency
$w^*$ that maximizes the marginal likelihood or minimizes the LOO
Error in the last run of \textbf{for} loop (lines 11-13).
\end{algorithmic}
%\end{algorithm}
\end{minipage}
}
\caption{Hyperparameter Optimization Algorithm}
\label{fig:alg}
\end{figure*}

We start by demonstrating experimentally that gradient based methods are not
sufficient for period estimation. We generate synthetic data and maximize
the marginal likelihood w.r.t. $\bm{\theta}=\{\beta, w, \ell\}$ using conjugate
gradients. For this experiment, 30 samples in the interval $[-10, 10]$ are
generated according to the periodic covariance function in Equation~(\ref{eqn:pk}) with $\bm{\theta}
= [1,0.25,1]$. Fixing $\beta, \ell$ to their correct values, the marginal
likelihood w.r.t. the period $1/w$ is shown in Figure~\ref{fig:1ab} left. The
figure shows that the marginal likelihood has numerous local minima in the high
frequency (small period) region that have no relation to the true period.
Figure~\ref{fig:1ab} right shows two functions with the learned parameters
based on different starting points (initial values).
%[PP: BW. Can you provide the values of the parameters ?
% I AM SURPRISED THE AMPLITUDE CHANGES
% STILL WOULD LIKE TO SEE THE NUMBERS
% BW: The value of the parameter is $\bm{\theta}$ in the earlier part of this paragraph.
% I think the idea we want to convey here is the oscillation, the exact number of the likelihood is not important.

 The function plotted in dark color
estimates the true function correctly while the one in light color
does not.  This is not surprising because from Figure~\ref{fig:1ab}
left, we can see that there is only a small region of initial points from
which the algorithm can find the correct period.
We repeated this experiment using
several other periodic functions with similar results.
These preliminary experiments illustrate two
points:
%(not shown here) of 1) a simple sine function; 2) a
%slightly more complex periodic function given by $y=\sin x + 0.7\sin
%2x - 0.2\cos 4x$.
\begin{itemize}
\item When other parameters are known, the marginal likelihood function is
  maximized at the correct period, showing that in principle we can find the
  correct period by optimizing the marginal likelihood.
\item On the other hand, it is not possible
to identify the period using only gradient based search.
%to optimize the marginal
%  likelihood function using gradient descent.
\end{itemize}

Therefore, as in previous work
\citep{reimann1994frequency,hall2000nonparametric}, our algorithm uses grid
search for the frequency. The grid used for the search must be
sufficiently fine
to detect the correct frequency and this implies high computational complexity.
We therefore follow a two level grid search for frequency where the coarse
grid must intersect the smooth region of the true maximum and the fine grid
can search for the maximum itself.
The two-level search significantly reduces the computational cost.
% and maintains high accuracy of the algorithm.
Our algorithm, presented in Figure~\ref{fig:alg} combines this with gradient based optimization of the
other parameters. There are several points that deserve further
discussion, as follows:

1. In step 3, we can successfully maximize the marginal likelihood w.r.t.\
$\beta, \ell$ and $\sigma^2$ using the conjugate gradients method, but this
approach does not work for the frequency $w$. The reason is that the
objective function is highly sensitive w.r.t.\ $w$ and the gradient is not
useful for finding the global maximum.
%[PP; YOU SAID THIS ABOVE]
This property justifies the structure
of our algorithm. This issues has been observed before and grid search (in
particular using two stages) is known to be the most effective
solution~\citep{reimann1994frequency,hall2000nonparametric}.

2. Our algorithm uses cyclic optimization estimating $w$, $\sigma$,
$\beta$, $\ell$. That is to say, we fix other parameters $\sigma$,
$\beta$, $\ell$ and optimize $w$ and then optimize $\sigma$,
$\beta$, $\ell$ when $w$ is fixed. We keep doing this iteratively but
use a small number of iterations (in our experiments, the default number of iterations is 2).
A more complete algorithm would
iterate until convergence but this incurs a large computational
cost. Our experiments demonstrate that a small number of iterations
is sufficient.

3. In steps 3 and 11 we incorporate $w$ into the joint optimization
of the marginal likelihood. This yields better results than optimizing
w.r.t.\ the other parameters with fixed $w$. This shows that the
gradient of $w$ sometimes still provides useful information locally,
although the obtained optimal value $\tilde{w}$ is discarded.

4. We use an adaptive search in the frequency domain, where at the
first stage we use a coarse grid and later a fine grid search is
performed at the neighbors of the best frequencies previously found.
By doing this, the computational cost is dramatically reduced while
the accuracy of the algorithm is still guaranteed.

Two additional approximations are introduced next,
specifically targeting the coarse and fine grids respectively and using
observations that are appropriate in each case.

\subsection{Ensemble Subsampling}

%In the coarse grid search of Algorithm 1, step 4 requires $|\mathcal{C}|$
%calculations of Equation~(\ref{eqn:malike}) which needs to compute the kernel
%matrix w.r.t.~each frequency in $\mathcal{C}$ and invert the corresponding
%kernel matrix. It takes $O(N^2)$ to calculate the kernel matrix, where $N$ is
%the number of samples of the time series. Note that in our astrophysics data,
%$N$ varies from star to star. The matrix inversion is normally implemented by
%Cholesky decomposition that requires $O(N^3)$ time, thus the total time
%complexity is $O\left(|\mathcal{C}|N^3\right)$. This greatly hampers the
%applicability of this algorithm when the coarse grid has a large cardinality.

The coarse grid search in lines 4-6 of the algorithm needs to
compute the covariance matrix w.r.t.~each frequency in $\mathcal{C}$ and
invert the corresponding covariance matrix, and therefore the total time
complexity is $\fO\left(|\mathcal{C}|N^3\right)$.  In addition,
different stars do not share the same
sampling points. Therefore the covariance matrix and its inverse cannot
be cached to be used on all stars.
 The computational cost is too
high when the coarse grid has a large cardinality. Our observation
here is that it might suffice to get an approximation of the
likelihood at this stage of the algorithm, because additional fine
grid search is done in the next stage.

Therefore, to reduce the time complexity, we propose an ensemble
approach that combines the marginal likelihood of several subsampled
times series.  The idea~\citep{protopapas2005fast} is that the
correct period will get a high score for all sub-samples, but wrong
periods that might score well on some sub-samples (and be preferred to
others due to outliers) will not score
well on all of them and will thus not be chosen. For the
approximation, we sub-sample the original time series such that it
only contains a fraction $f$ of the original time points, repeating
the process $R$ times.  The marginal likelihood score is the average
over the $R$ repetitions.  Our experiments
%explore different values for $f$ and $R$ and
justify default settings of $f=15\%$ (with the additional constraint
that $30\leq f\leq 40$) and $R=10$.
%[PP: WEIRD CONSTRAINT] % BW: Because we don't want the subsamples to be too small or too large.
This approximation reduces the
time complexity to $\fO\left(|\mathcal{C}|\times R \times
(fN)^3\right)$.

\subsection{First Order Approximation with Low Rank Approximation}
Similar to the previous case, the time complexity of fine grid
search is $\fO(|\mathcal{F}|N^3)$. In this case we can reduce the
constant factor in the $\fO(N^3)$ term. Notice that in step 13,
other parameters are fixed and the grid is fine so that the marginal
likelihood is a smooth function of $w$. Suppose we have $w_0,
w_1\in\mathcal{F}$ where $\cal{F}$ is the fine grid
and $\Delta w = |w_0 - w_1| < \epsilon$, where
$\epsilon$ is a predefined threshold. Then, given $\K_{w_0}$, the
covariance matrix w.r.t. $w_0$, we can get $\K_{w_1}$ by its Taylor
expansion as
\begin{equation}
  \K_{w_1} = \K_{w_0} + \frac{\partial\K}{\partial w}(w_0)\Delta w + o(\epsilon^2).
\end{equation}
Denote $\widetilde{\K} = \frac{\partial\K}{\partial w}(w_0)$ where
$\widetilde{\K}\Delta w$ can be seen as a small perturbation to
$\K_{w_0}$. At first look, the Sherman-Morrison-Woodbury
formula~\citep{bishop2006pattern} appears to be suitable for
calculating the
update of the inverse efficiently. Unfortunately, preliminary experiments (not
shown here) indicated that this method fails due to numeric
instability. Instead, we use an update for the Cholesky
factors of the matrix and calculate the inverse through these.
Namely, given the Cholesky decomposition of
$\K_{w_0}=\bL\bL^T$ we calculate $\widetilde{\bL}$ such that
$\widetilde{\bL}\widetilde{\bL}^T=\K_{w_0}+\Delta
w\widetilde{\K}\approx \K_{w_1}$.
Details of this construction are given
in the appendix.

\subsection{Astrophysical Input Improvements}

For some cases we may have further information on the type of
periodic functions one might expect. We propose to use such
information to bias the selection of periods, by using it to induce
a prior over periods or as a post-processing selection criterion.
The details of these steps are provided in the next section.

%% experimental part
\section{Experiments}
This section evaluates the various algorithmic ideas using synthetic
and astrophysics data and then applies the algorithm to a different set
of lightcurves. Our implementation of the algorithms makes use of the
{\em gpml} package~\citep{rasmussen2010gaussian}\footnote{http://www.gaussianprocess.org/gpml/code/matlab/doc/}.

\subsection{Synthetic data}
%\subsubsection{Justifying grid search}
%Preliminary experiments (not shown here) demonstrate that for the data sampled from a GP, when other
%hyperparameters are known, the marginal likelihood function is maximized at the
%correct period. This shows that technically we can find the correct period by
%optimizing the marginal likelihood. However, it is not possible to
%optimize the marginal likelihood function via gradient descent due to
%high sensitivity and numerous local maxima when
%%the high frequency oscillation when
%the period is small.
%Therefore, we come to the conclusion that a grid search is necessary
%to find the period that optimizes the likelihood.

%\subsubsection{Comparisons under different scenarios}

In this section, we evaluate the performance of several variants of our algorithm, study the effects of its parameters, and compare it to the two most used methods in the literature: the LS periodogram
(LS)~\citep{lomb1976least} and phase dispersion minimization
(PDM)~\citep{stellingwerf1978period}.

%Suppose $\{x_j,y_j\},
%j=1,\cdots,N$ are samples and observations respectively,
The LS method \citep{lomb1976least} chooses $w$ to maximize
%Equation~(\ref{eqn:LS}).
the periodogram defined as:
\begin{equation}
\label{eqn:LS}
\begin{split}
P_{LS}(\omega) &= \frac{1}{2}\Bigg\{\frac{[\sum
y_j\cos(\eta_j)]^2}{\sum\cos^2(\eta_j)}+
\frac{[\sum
y_j\sin(\eta_j)]^2}{\sum\sin^2(\eta_j)}\Bigg\},
\end{split}
\end{equation}
where $\eta_j = \omega(x_j-\tau)$. The phase $\tau$ (that depends on
$\omega$) is defined as the value satisfying $ \tan(2\omega\tau) =
\frac{\sum\sin(2\omega x_j)}{\sum\cos(2\omega x_j)}.  $ As shown
by~\citep{reimann1994frequency}, LS fits the data with a harmonic model using
least-squares.

In the PDM method, the period producing the least possible scatter in the
derived light curve is chosen.  The score for a proposed period can be
calculated by folding the light curve using the proposed period, dividing the
resulting observation phases into bins, and calculating the local variance
within each bin, $ \sigma^2 = \frac{\sum_j(y_j - \bar{y})^2}{N - 1}, $ where
$\bar{y}$ is the mean value within the bin and the bin has $N$ samples. The
total score is the sum of variances over all the bins. This method has no
preference for a particular shape (e.g., sinusoidal) for the curve.

\begin{figure*}[t]
{
        \begin{minipage}{0.45\textwidth}
        \centering
        \includegraphics[width=\textwidth]{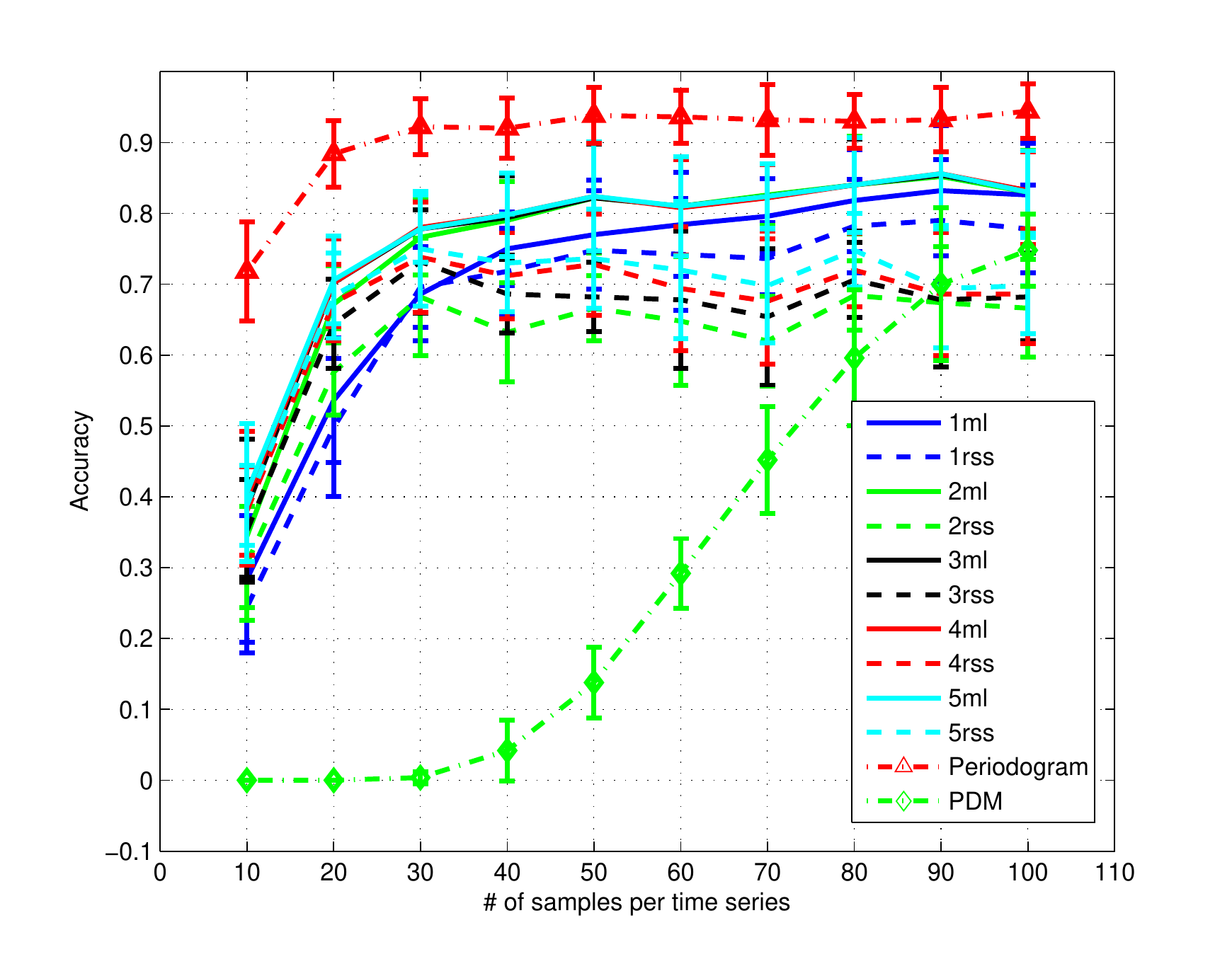}
        \end{minipage}%\hfill
        \begin{minipage}{0.45\textwidth}
        \centering
        \includegraphics[width=\textwidth]{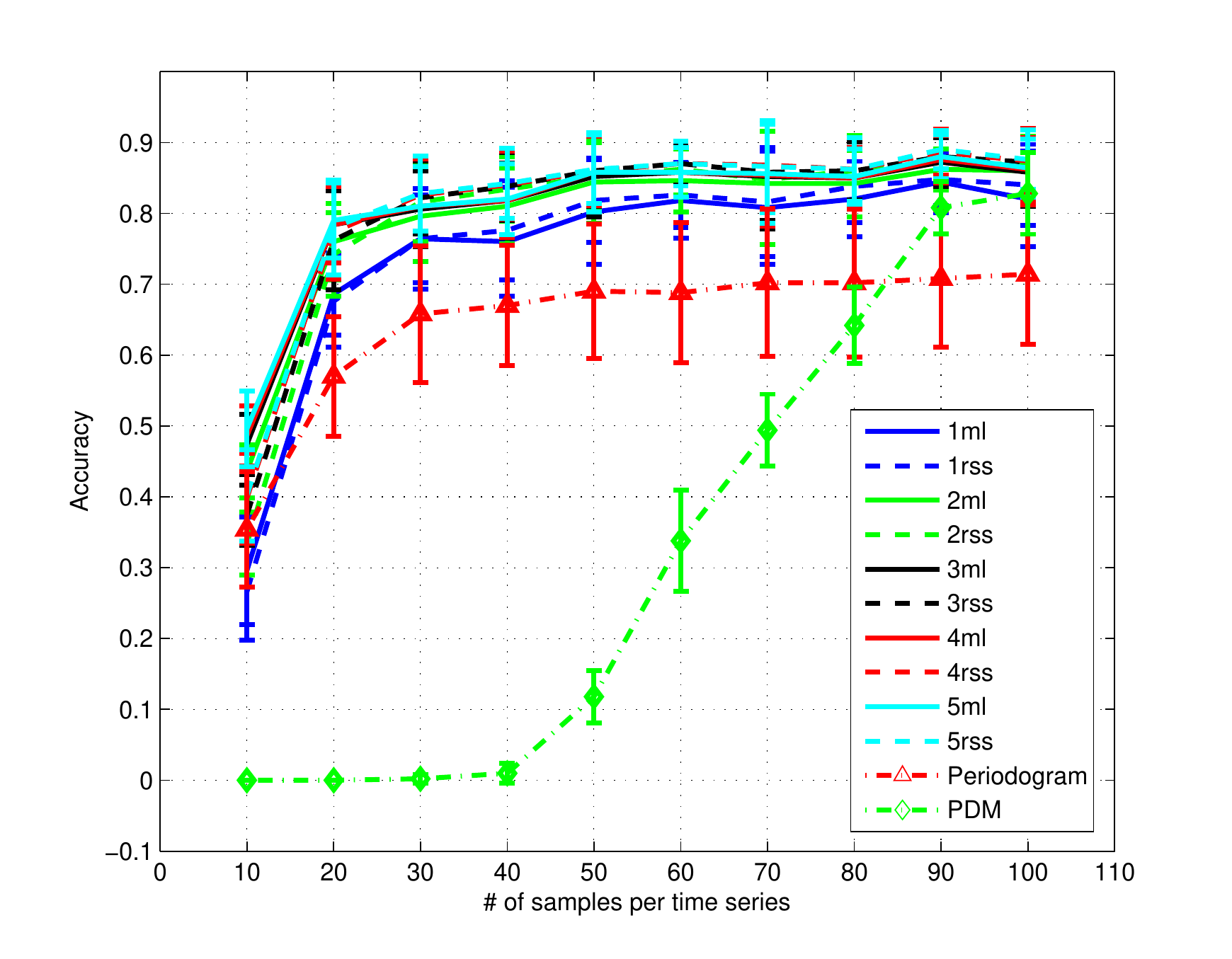}
        \end{minipage}%\hfill

        \begin{minipage}{0.45\textwidth}
        \centering
        \includegraphics[width=\textwidth]{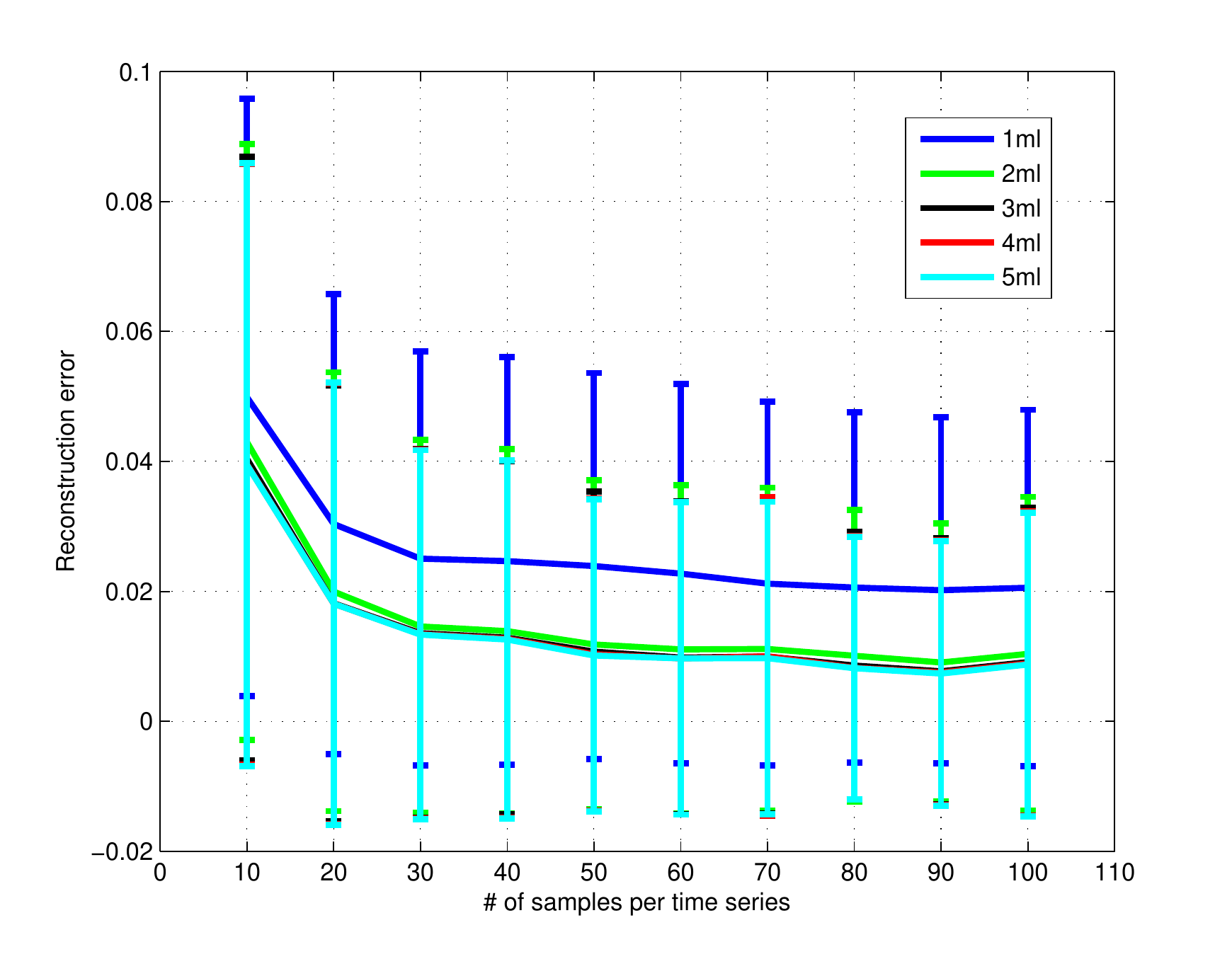}
        \end{minipage}
        \begin{minipage}{0.45\textwidth}
        \centering
        \includegraphics[width=\textwidth]{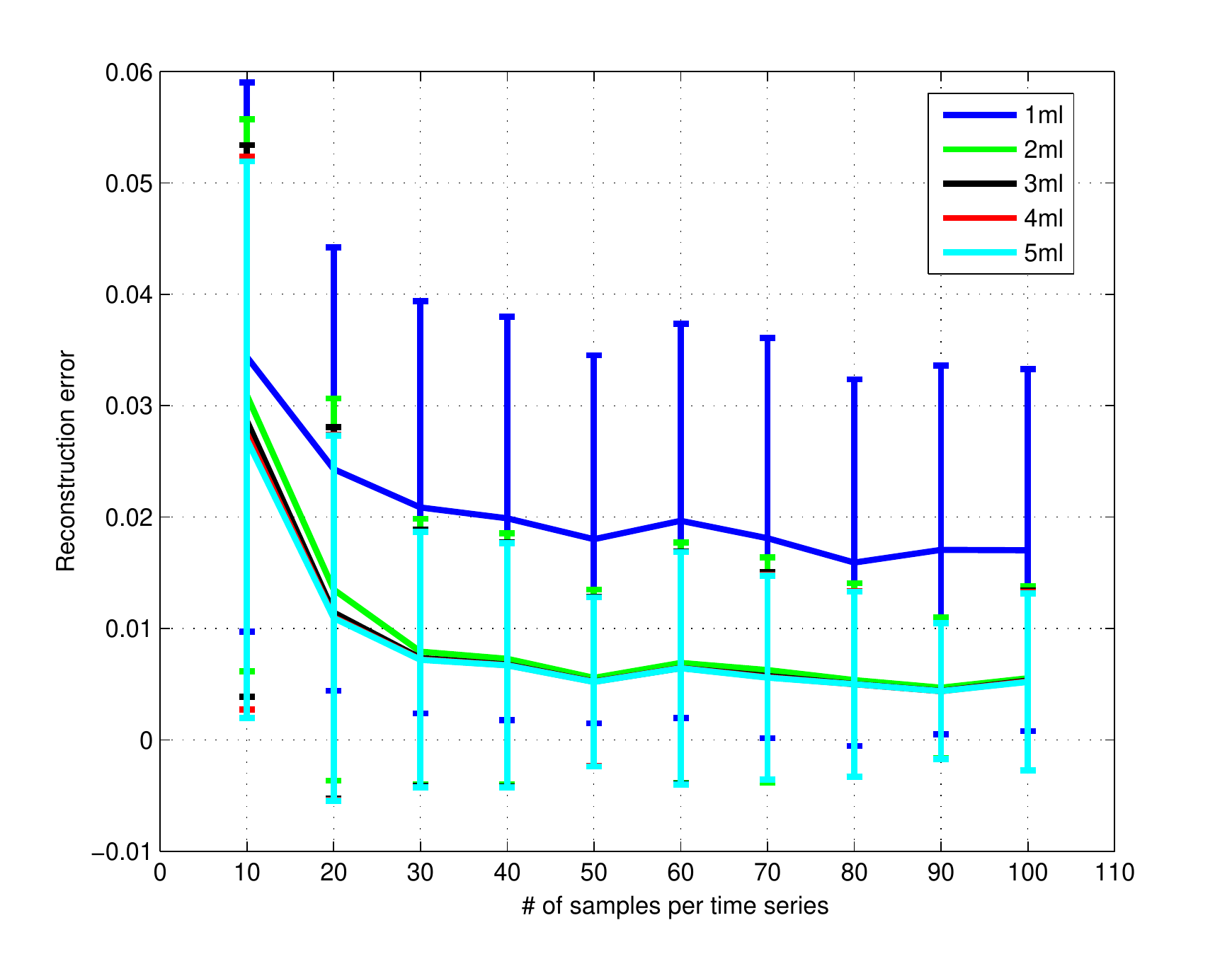}
        \end{minipage}

        \begin{minipage}{0.45\textwidth}
        \centering
        \includegraphics[width=\textwidth]{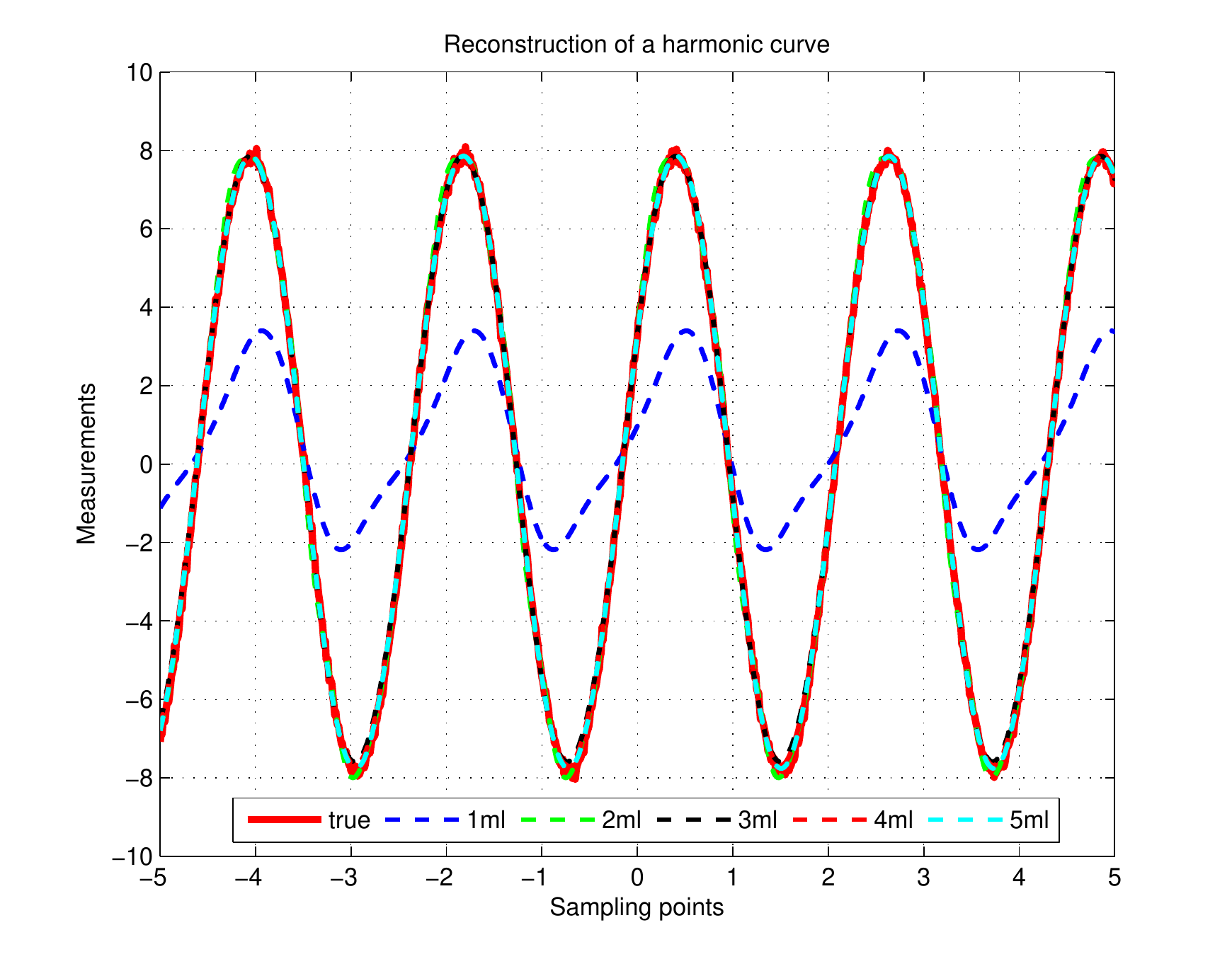}
        \end{minipage}%
        \begin{minipage}{0.45\textwidth}
        \centering
        \includegraphics[width=\textwidth]{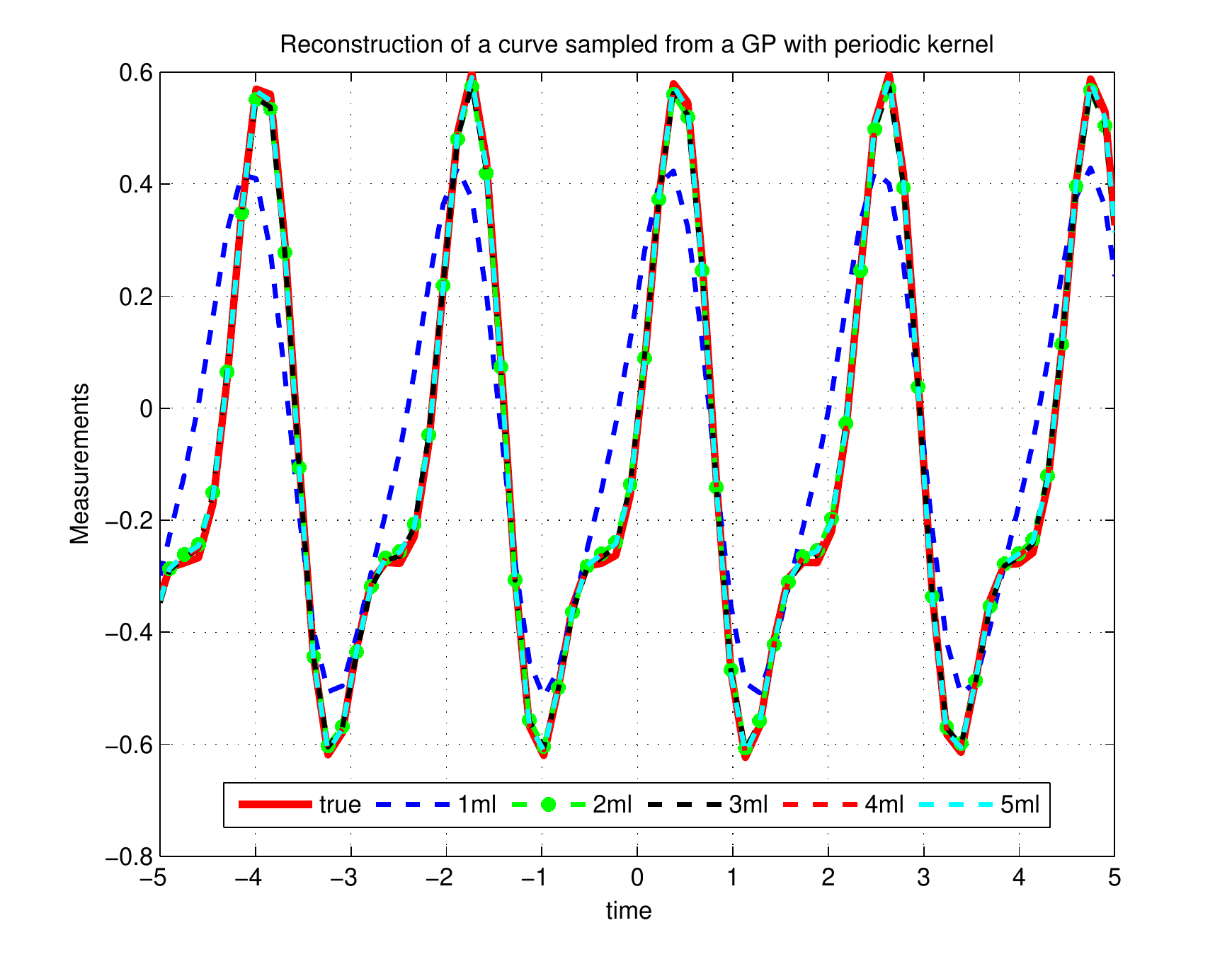}
        \end{minipage}
\caption{
Results for harmonic data (left column) and GP data (right column).
Left: Accuracy (mean and standard deviation) versus the
number of samples, where solid lines marked with $n$ml
represent GP with marginal likelihood where $n$ denotes the number of
iterations. The corresponding dotted lines marked $n$rss denote cross-validation
results with $n$ iterations. Middle: Reconstruction error for the regression function versus the number of samples.
%Results are from 10 repetitions each representing 50
%random functions.
Right: Reconstruction curve of GP in two specific runs using maximum
likelihood with
different number of iterations.}
%\label{fig:gpsine}}
\label{fig:artdataplots}}
\end{figure*}

We generate two types of artificial data, referred to as harmonic data and GP
data below.  For the first, data is sampled from a simple harmonic function,
\begin{equation}
y \sim \mathcal{N}\left(a\sin(\omega x + \phi_1) + b\cos(\omega x +
\phi_2), \sigma^2\mathbb{I}\right)
\end{equation}
where $a,b\sim\text{Uniform}(0,5)$, $\omega\sim\text{Uniform}(1,4)$,
$\phi_i\sim\mathcal{N}(0,1)$ and the noise level $\sigma^2$ is set to be
0.1.
Note that this is the model assumed by LS. For the second, data is
sampled from a GP with periodic covariance function in Equation~(\ref{eqn:pk}). We generate
$\beta, \ell$ uniformly in $(0,3]$ and $(0,3]$ respectively and the noise
level $\sigma^2$ is set to be 0.1. The period is drawn from a uniform
distribution between $(0.5, 2.5]$.
For each type we generate
data under the following configuration. We randomly sampled 50 time
series each having 100 time samples
in the interval $[-5,5]$. Then the comparison is
performed using sub-samples with size increasing from 10 to 100.
This is repeated ten times to generate means and standard deviations in the
plots.

The setting of the algorithms is as follows:  In our algorithm we only use
one stage grid search.  For our algorithm and LS, the lowest
frequency $f_{\min}$ to be examined is the inverse of the span of the input
data $1/(x_{\max}-x_{\min})=1/T$. The highest frequency $f_{\max}$ is twice
the Nyquist frequency $f_N$, which we would obtain, if the data points were
evenly spaced over the same span $T$, that is $f_N = N/(2T)$. We use an
over-sample factor of 8, meaning that the range of frequencies is broken into
even segments of $1/8T$. For PDM we set the frequency range to be
$[0.02, 5]$ with the frequency increments of 0.001 and the number of bins in
the folded period is set to be 15.

For performance measures we consider both ``accuracy'' in identifying the
period and the error of the regression function. For accuracy, we consider an
algorithm to correctly find the period if its error is less than $1\%$ of the
true period, i.e., $|\hat{p} - p|/p\leqslant 1\%$. Further experiments (not
shown here) justify this approach by showing that the accuracies reported are
not sensitive to the predefined error threshold.

The results, where our algorithm does not use the sampling and low rank
approximations, are shown in Figure~\ref{fig:artdataplots} and they support
the following observations.

1. As expected, the top left plot shows that LS performs very well on the
harmonic data and it outperforms both PDM and our algorithm.
This means that if we know that the expected shape is sinusoidal,
then LS is the best choice. This confirms the conclusion of other
studies. For example, in the problem of detecting periodic genes from
irregularly sampled gene expressions~\citep{wentao2008detecting,
  glynn2006detecting}, the periodic time series
of interest
%for which the biologists were looking
were exactly sine curves. In this case, studies showed that LS is the
most effective comparing to several other statistical models.
%[PP: WE NEED ASTROPHYSICAL JOURNAL EXAMPLES -FOR PP]

2. On the other hand, the top right plot shows that our algorithm is
significantly better than LS on the GP data showing that when the curves are
non-sinusoidal the new model is indeed useful.

%As discussed above, one can extend Algorithm 1 using a cyclic optimization
%over $w$ and $(\beta, \ell, \sigma)$ until it converges.
%However, Figure~(\ref{fig:artdataplots})

3.  The two plots in
top row together show that our algorithm performs significantly better
than PDM on
both types of data, especially when the number of samples is small.

4. The first two rows show the performance of the cyclic optimization
procedure with 1-5 iterations. We clearly see that for these datasets there
is little improvement beyond two iterations.  The bottom row shows two
examples of the learned regression curves using our method with different number of
iterations. Although one iteration does find the correct period, the
reconstruction curves are not accurate.  However, here too, there is
little improvement beyond two iterations.  This shows that for the
data tested here two iterations
suffice for period estimation and for the regression problem.

5. The performance of marginal likelihood and cross validation is close, with
marginal likelihood dominating on the harmonic data and doing slightly worse
in GP data.

%4. From the left part of Figure~(\ref{fig:artdataplots}), we can see that as
%the number of samples increase, the accuracy gradually increases, which can
%be seen as a preliminary result toward showing the consistency of our period
%estimation.

%[PP. FONT SIZES HAVE TO INCREASE ]
\begin{figure*}[t]
{
        \begin{minipage}{0.33\textwidth}
        \centering
        \includegraphics[width=\textwidth]{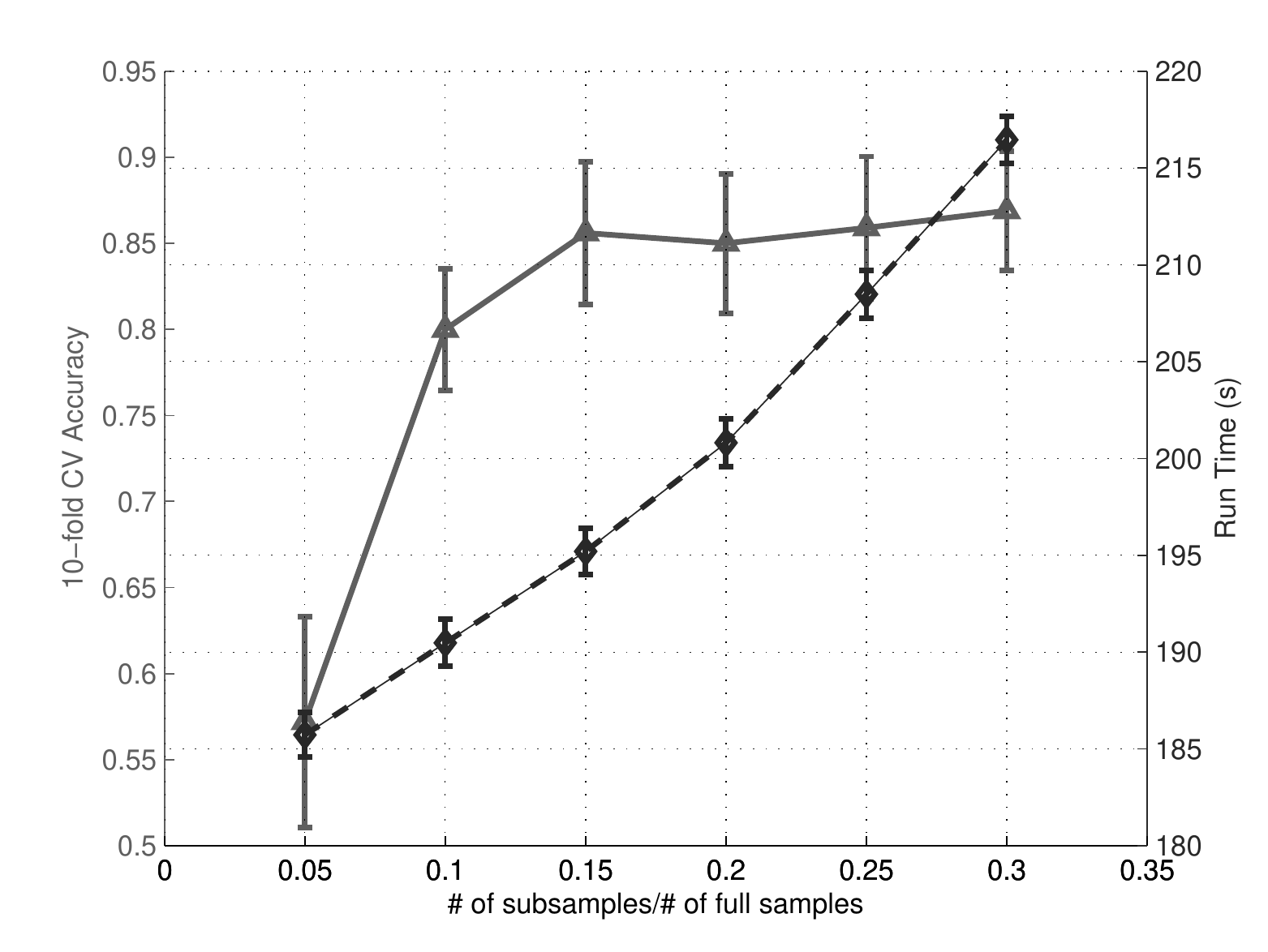}
        \end{minipage}%\hfill
        \begin{minipage}{0.33\textwidth}
        \centering
        \includegraphics[width=\textwidth]{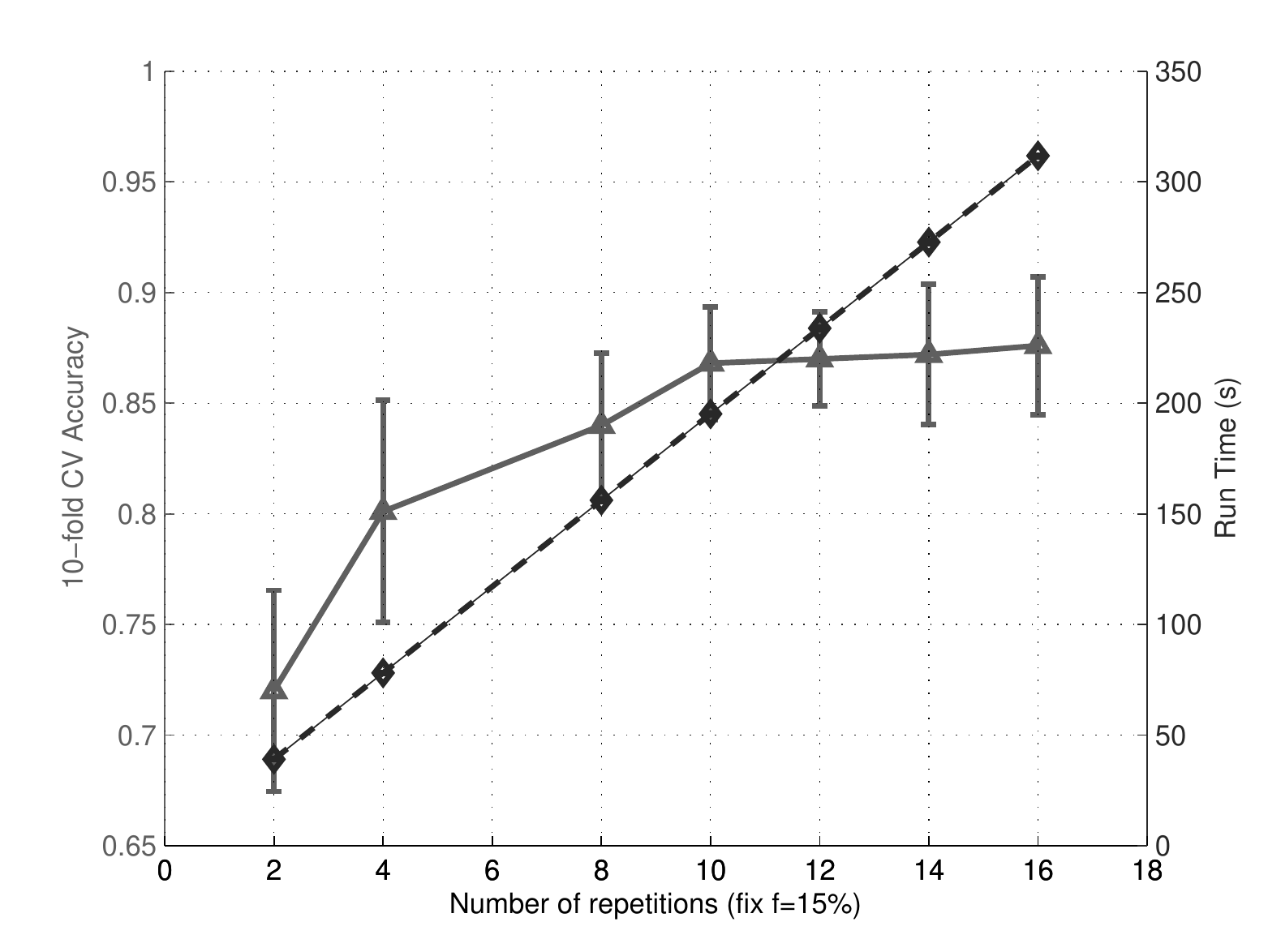}
        \end{minipage}
        \begin{minipage}{0.33\textwidth}
        \centering
        \includegraphics[width=\textwidth]{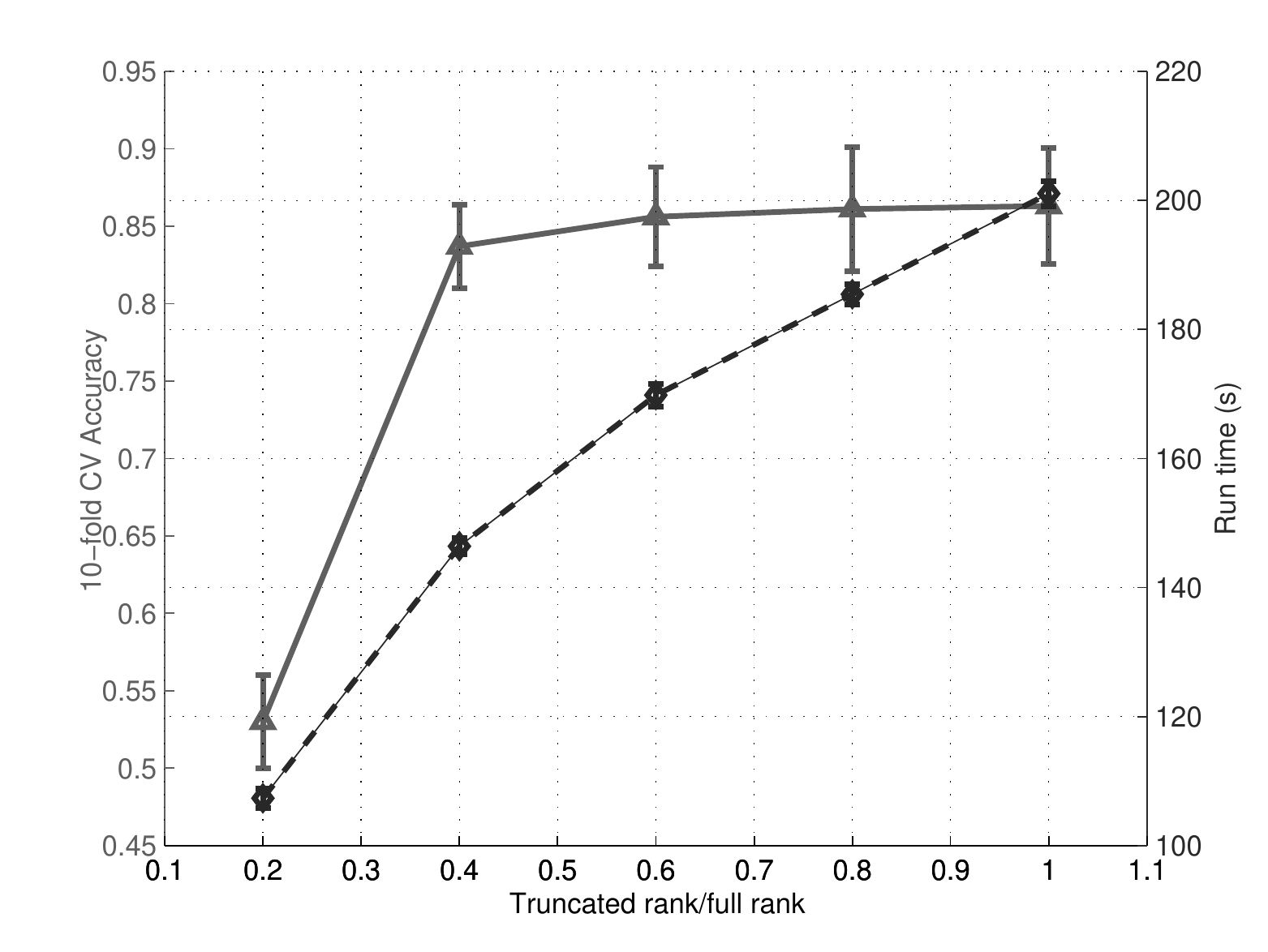}
        \end{minipage}
\caption{Accuracy (solid line) and Run time (dash-line) of approximation methods as a function of
  their parameters.
Left: sub-sampling ratio (with $R=10$).
Middle: number of repetitions (with 15\% sub-sampling).
Right: rank in low rank approximation.
}
\label{fig:gpsub}}
\end{figure*}

We next investigate the performance of the speedup techniques. For this we
use GP data
%with 100 samples
under the same configuration as the previous
experiments.  The experiment was repeated 10 times where in each round
we generate 100 lightcurves
each having 100 samples but generated from different $\theta$s.  For the
algorithm we used two iterations for cyclic optimization and varied the
subsampling size, number of repetitions and rank of the approximation.
Table~\ref{tab:time} shows results with our chosen parameter setting using
sampling rate of 15\%, 10 repetitions, approximation rank
$M=\lfloor\frac{N}{2}\rfloor$ and grid search threshold $\epsilon = 0.005$.
We can see that the subsampling technique saves over 60\% percent of the run
time while at the same time slightly increasing the accuracy. Low rank
Cholesky approximation leads to an additional 15\% decrease in run time, but
gives slightly less good performance.
% due to the reduced rank approximation.
Figure~\ref{fig:gpsub} plots the performance of the speedup methods under
different parameter settings.  The figure clearly shows that the chosen
setting provides a good tradeoff in terms of performance vs.\ run time.

%We also compare their performances under different settings, as shown
%in Figure~(\ref{fig:gpsub}).
%From the results, it can be seen that $15\%$ of the original samples is
%sufficiently good for period estimation. Regarding low rank Cholesky update,
%if we do a rough approximation at around 20\% of the full rank, we can't get
%good performance. While the truncated rank increase to around half, we get
%reasonably good performance while it costs much less than full
%decomposition. Fixing number of subsamples to be 15, we can see that
%increasing the number of repetition do improve the performance. However, the
%increase computational cost outweighs the benefit of the performance
%improvement.
%

%a $15\%$ fraction of the original samples is sufficiently good for period
%estimation.  Fixing number of subsamples to be 15, we can see that increasing
%the number of repetitions does improve the performance. However, the increase
%computational cost outweighs the benefit of the performance improvement.  For
%the low-rank approximation, we can see that a rough approximation say at 20\%
%of the full rank degrades performance significantly. At 50\% of the rank the
%accuracy is maintained while providing some speedup.

  \begin{table}[t]
    \caption{Comparison of GPs: Original, Subsampling and Subsampling
    plus low rank Cholesky update. \textsc{Acc} denotes accuracy and \textsc{S/TS} denotes the running time in seconds per time series.
    %[PP: IF WE REPORT SECONDS THEN YOU MUST REPORT MACHINE AND LANGUAGE.
    % SAME CODE IN C COULD BE 10 TIMES FASTER
    % BW: Fair enough, we can add this.
    }
\centering
    \begin{tabular}{||c||c|c|c||}
    \hline
    \textsc{} &  \textsc{original} & \textsc{subsampling} & \textsc{sub + lowR} \\
    \hline
    \textsc{Acc} & $0.831\pm 0.033$ & $\mathbf{0.857\pm 0.038}$  &   $0.849\pm 0.028$ \\
    \hline
    \textsc{s/ts} & $518.52\pm 121.49$ & $197.59\pm 14.10$
    & $\mathbf{170.75\pm17.93}$\\
    \hline
    \end{tabular}
    \label{tab:time}
    \end{table}

% \begin{figure*}[t]
% {
%         \begin{minipage}{0.33\textwidth}
%         \centering
%         \includegraphics[width=\textwidth]{fig/gp_gp1.eps}
%         \end{minipage}%\hfill
%         \begin{minipage}{0.33\textwidth}
%         \centering
%         \includegraphics[width=\textwidth]{fig/gp_gp2.eps}
%         \end{minipage}
%         \begin{minipage}{0.33\textwidth}
%         \centering
%         \includegraphics[width=\textwidth]{fig/demo_gp.eps}
%         \end{minipage}
% \caption{Left: Accuracy (mean and standard deviation) versus the
% number of samples for GP data, where solid lines with $n$ml mean GP
% with marginal likelihood where $n$ denotes the number of iterations.
% The corresponding rss denotes the Cross-validation version. Middle:
% Reconstruction versus the number of samples for GP data. Results are
% from 10 repetitions each representing 50 random
% functions. Right: Reconstruction curve (one example) of GP using Maximum likelihood with
% different iterations on GP data. }\label{fig:gpgp}}
% \end{figure*}

%% OGLE exp

\subsection{Astrophysics Data}

 In this section, we estimate the
periods of unfolded astrophysics time series from the OGLEII survey
\citep{soszynski2003optical}.

\begin{figure*}[t]
{
        \begin{minipage}{0.33\textwidth}
        \centering
        \includegraphics[width=\textwidth]{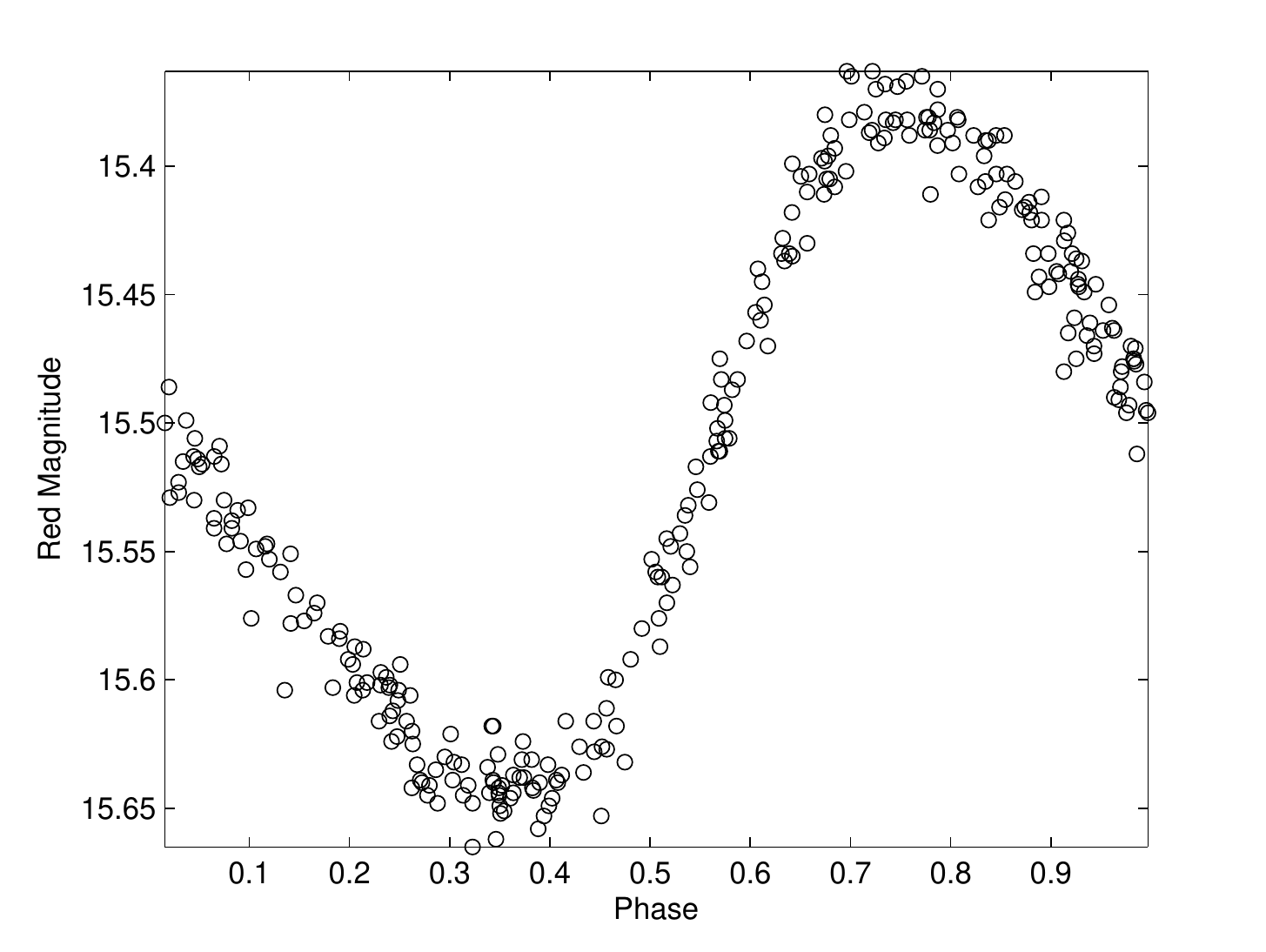}
        \end{minipage}%\hfill
        \begin{minipage}{0.33\textwidth}
        \centering
        \includegraphics[width=\textwidth]{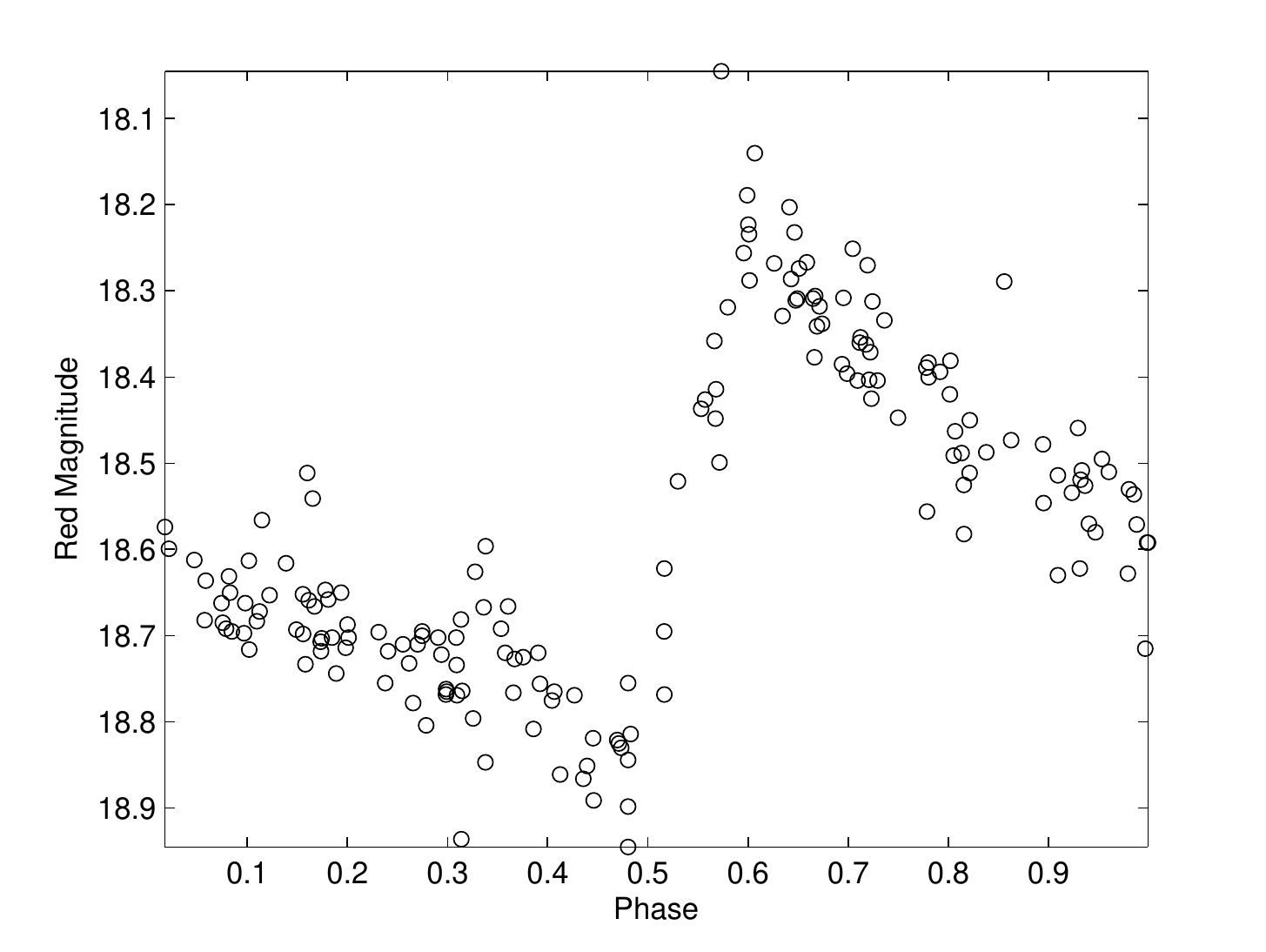}
        \end{minipage}
        \begin{minipage}{0.33\textwidth}
        \centering
        \includegraphics[width=\textwidth]{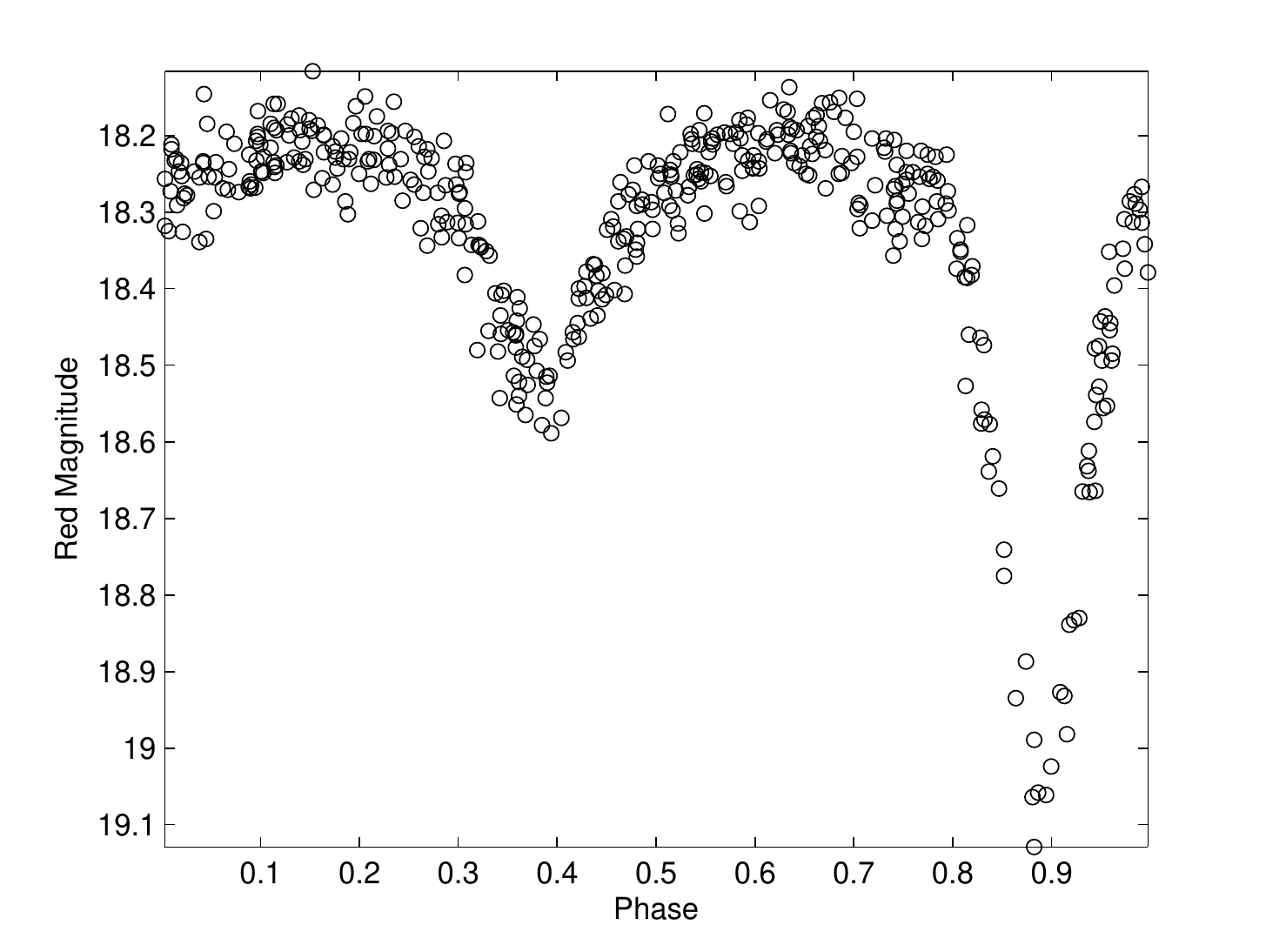}
        \end{minipage}
\caption{Examples of light curves of periodic variable stars folded
according to their period to highlight the periodic shape.
%Each column shows the folded stars of different types.
Left: Cepheid, middle: RR Lyrae, right: Eclipsing Binary.}
\label{typical}}
\end{figure*}

OGLE surveyed the sky over a number of years and has a huge number
of light sources. The data we use here is a subset of OGLEII,
containing a total of 14087 light curves of periodic variable stars
that have previously been identified to be periodic (and thus their
period is known) and to be members of one of 3 types: Cepheids, RR
Lyrae, and Eclipsing Binary (illustrated in Figure~\ref{typical}).

We first explore, validate and develop our algorithm using a subset
of OGLEII data and then apply the algorithm to the full OGLEII data\footnote{http://www.cs.tufts.edu/research/ml/index.php?op=data\_software}
except this development set. The OGLE subset is chosen to have 600
time series in total where each category is sampled according to its
proportion in the full dataset.

\subsubsection{Evaluating the General GP Algorithm}
The setting for our algorithm is as follows: The grid search ranges are
chosen to be appropriate for the application using coarse grid of $[0.02, 5]$
in the frequency domain with the increments of 0.001.  The fine grid is a
0.001 neighborhood of the top frequencies each having 20 points with a step
of 0.0001.  We use $K=20$ top frequencies in step~9 of the algorithm and vary
the number of iterations in a cyclic optimization.  When using
sub-sampling, we use 15\% of the original time series, but restrict sample
size to be between 30 and 40 samples. This guarantees that we do not use too
small a sample and that complexity is not too high.
For LS we use the same configuration as in the synthetic
experiment. Results are shown in Table~\ref{tab:sub1} and they mostly
confirm our conclusions from the synthetic data.
In particular, ML is slightly
better than CV and subsampling yields a small improvement.
In contrast with the artificial data,
%in this case
%In this case we see that
more iterations do provide a small improvement in
performances and 5 iterations provide the best results in this
experiment. Finally, we can also see that all of the GP variants outperform
LS.

\begin{table}[t]
\caption{Comparisons of different GPs on
OGLEII subset. \textsc{gp-ml} and \textsc{gp-cv} are GP with the ML and CV criteria.
\textsc{sgp-ml} and \textsc{sgp-cv} are the corresponding
subsampling versions.
% of GP with ML and CV.
The first column denotes the number of iterations.
%used in the optimization.
}
\centering
\begin{tabular}{||c||c|c|c|c|c||}
\hline
&  \textsc{gp-ml} & \textsc{gp-cv} & \textsc{sgp-ml} & \textsc{sgp-cv} &\textsc{ls}\\
\hline
\textsc{1itr acc} & $0.7856$ & $0.7769$ & $0.7874$ & 0.7808 & 0.7333\\
\hline
\textsc{2itr acc} & $0.7892$ & $0.7805$ & $0.7910$ & 0.7818 & -\\
\hline
\textsc{3itr acc} & $0.7928$ & $0.7806$ & $0.7964$ & 0.7845 & -\\
\hline
\textsc{4itr acc} & $0.7946$ & $0.7812$ & $0.7982$ & 0.7875 & -\\
\hline
\textsc{5itr acc} & $0.7964$ & $0.7823$ & $0.8000$ & 0.7906 & -\\
\hline
\end{tabular}
\label{tab:sub1}
\end{table}

%This  clearly shows an improvement over  LS but an
%accuracy of 80\% is still not satisfactory.
Although this is an improvement over existing algorithms
accuracy of 80\% is still not satisfactory.
As discussed by
~\cite{wachman2009thesis}, one particularly challenging task is
finding the true period of EB stars. The difficulty comes from the following
two aspects. First, for a symmetric EB, the true period and half of the true
period are not clearly distinguishable quantitatively. Secondly, methods that
are better able to identify the true period of EBs are prone to find periods
that are integer multiples of single bump stars like RRLs and Cepheids. On
the other hand, methods that fold RRLs and Cepheids correctly often give ``half''
of the true period of EBs. In particular, the low performance of LS is
due to the fact that it gives a half or otherwise wrong period for most
EBs.

\begin{figure*}[t]
{
        \begin{minipage}{0.49\textwidth}
        \centering
        \includegraphics[width=\textwidth]{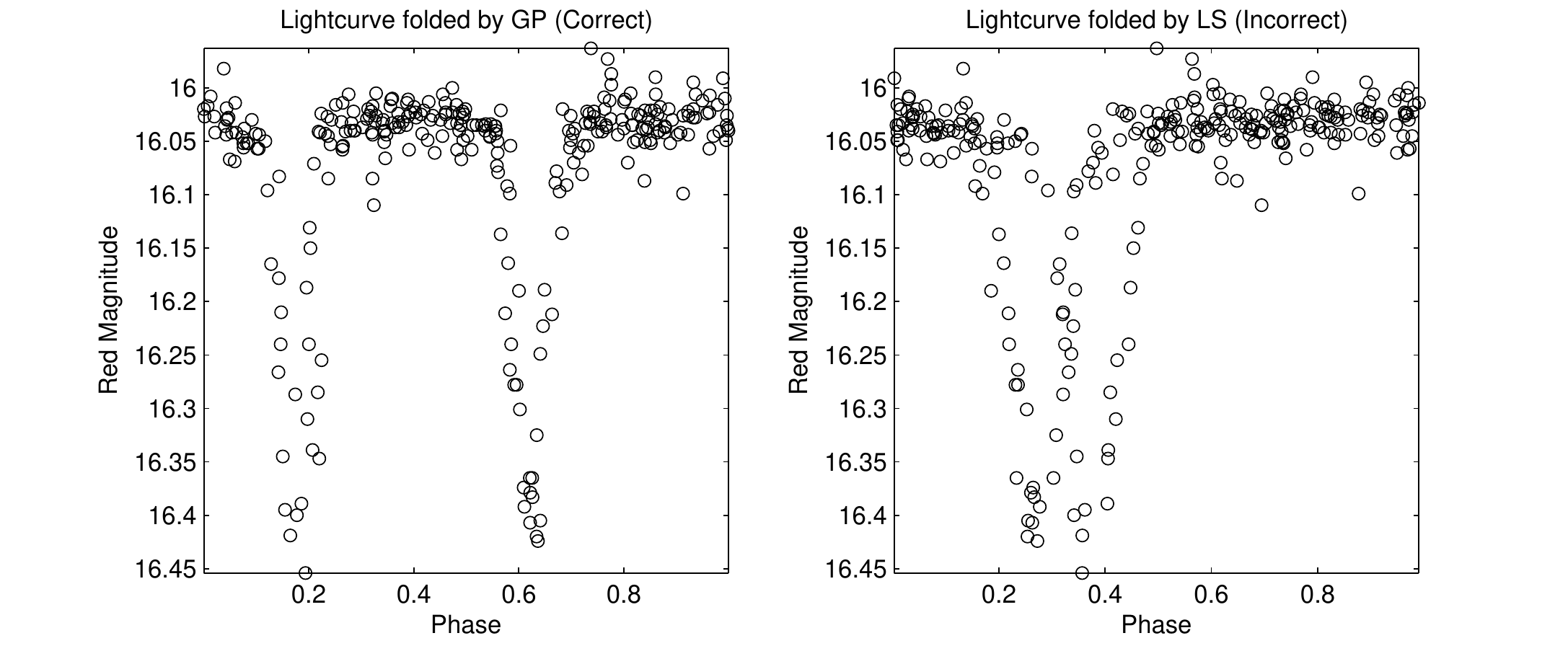}
        \end{minipage}%\hfill
        \begin{minipage}{0.49\textwidth}
        \centering
        \includegraphics[width=\textwidth]{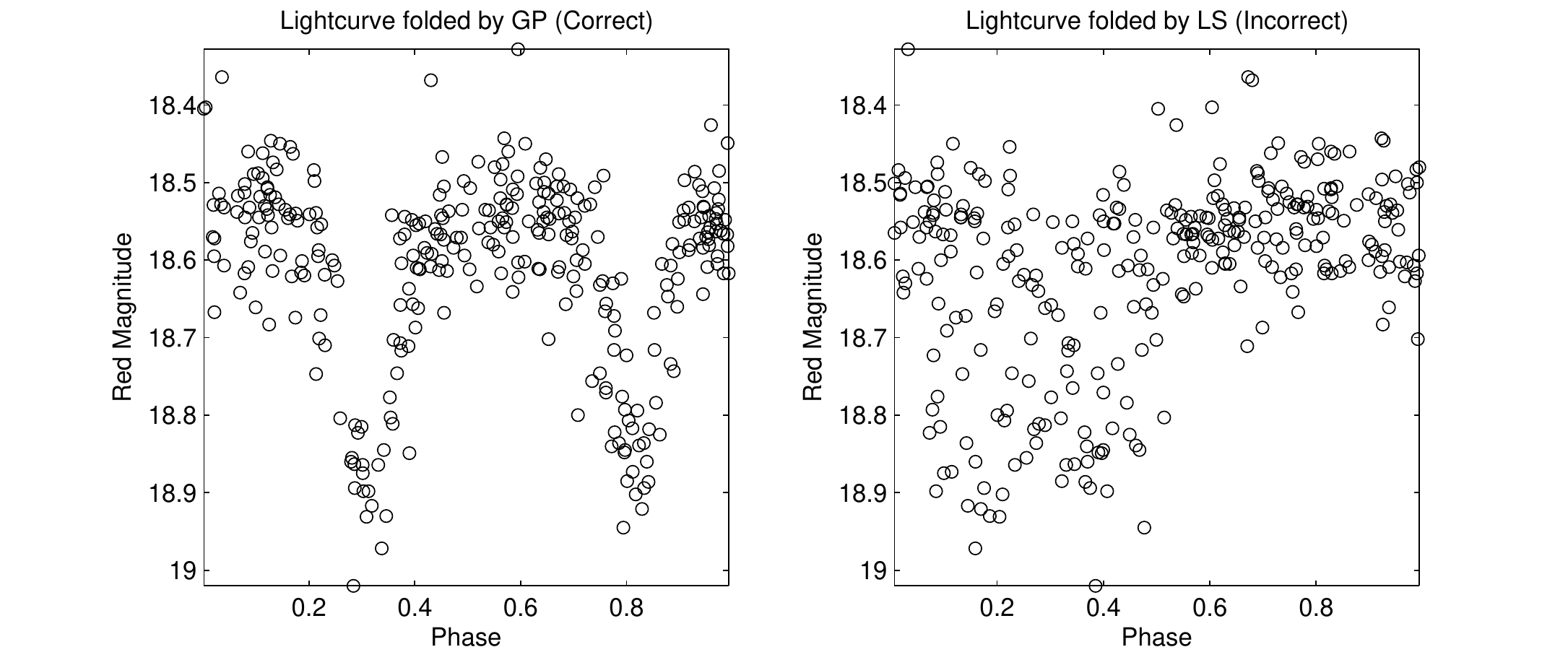}
        \end{minipage}
        \begin{minipage}{0.49\textwidth}
        \centering
        \includegraphics[width=\textwidth]{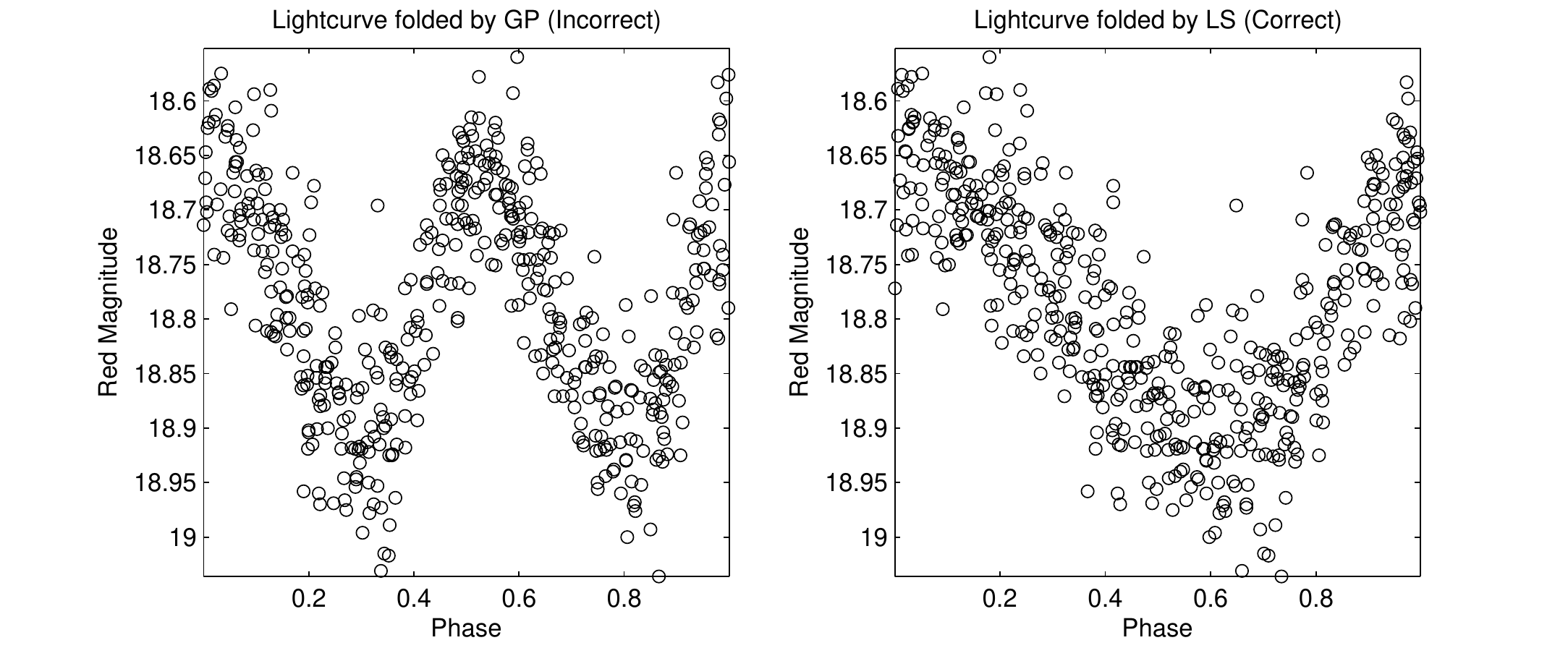}
        \end{minipage}
       \begin{minipage}{0.49\textwidth}
        \centering
        \includegraphics[width=\textwidth]{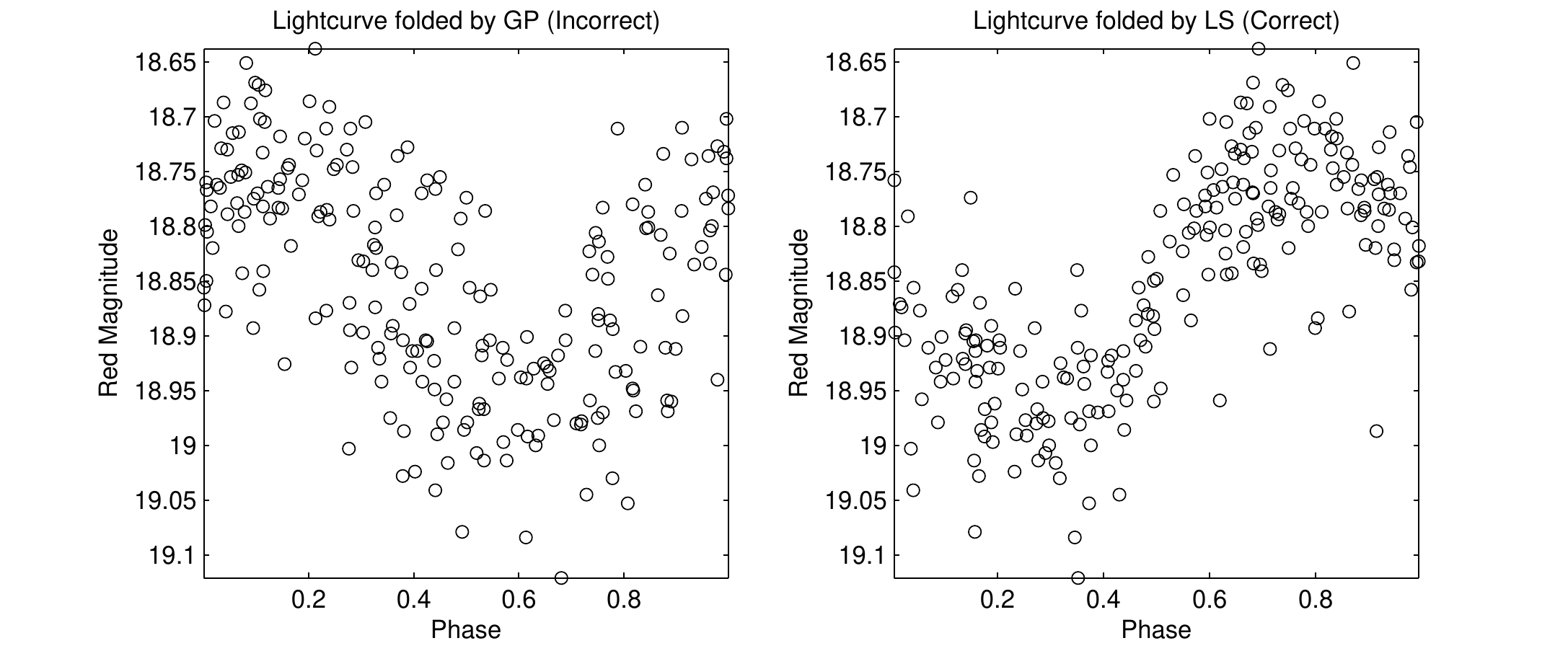}
        \end{minipage}
\caption{Examples of light curves where GP and LS and identify different periods and one of them is correct.
Each pair shows the time series folded by GP on the left and LS on the right. The top row shows cases where LS identifies half the period. The bottom row shows cases where GP identifies double the period or a different period.}
\label{fig:successandfail}}
\end{figure*}

To illustrate the results Figure~\ref{fig:successandfail} shows the periods
found by our method and by GP on 4 stars. The top row shows 2 cases
where the GP method finds the correct period and LS finds half the period.
The bottom row shows cases where LS identifies the correct
period and the GP does not. In the example on the left the GP doubles
the period. In the example on the right the GP identifies a different period
from LS but given the spread in the correct period the period it uncovers
is not unreasonable.

%\subsubsection{Astrophysical Dependent Improvements}
\subsubsection{Incorporating Domain Knowledge}

We next show how this issue can be alleviated and the performance
can be improved significantly using a learned probabilistic
generative model.
The methods developed
are general
and can be applied whenever such a model is available.
As illustrated in Figure~\ref{typical}, our astrophysics
knowledge suggests that different types of stars have different
typical shift-invariant ``shapes''.
In addition,
each class has more than one such shape and
each individual star has
some variation from the common shape. We use the
\emph{Shift-invariant Grouped Mixed-effect Model}
(\textsc{gmt})~\citep{wang2010shift}, which captures the common
``shapes'' via a mixture of Gaussian processes while at the same
time allowing for individual variations. This model was previously
developed to capture and aid in the classification of the
astrophysics data.
%The model parameters are learned via \textsc{em}
%algorithm using the training data; we refer interested readers
%to~\cite{wang2010shift}.
Once model parameters are learned
%a generative model for each
%type of stars from which
we can calculate the likelihood of a light
curve folded using a proposed period.
Given the models,
learned from a disjoint set of time series, for Cepheids, EBs and
RRLs with parameter sets $\mathcal{M}_i, i=\{C,E,R\}$, there are two
perspectives on how they can be used:

    \begin{table*}[t]
        \caption{Comparison of different regularization parameters on OGLEII subset using MAP.}
        \centering
        \begin{tabular}{||c||c|c|c|c|c|c|c||}
        \hline
        $\gamma$ &  \textsc{0} & \textsc{.1} & \textsc{.3} & \textsc{.5} & \textsc{.7} & \textsc{.9} & 1\\
        \hline
        \textsc{acc} & 0.87027  & 0.85946  &   0.81802  &   0.81802  &   0.80901  &   0.80721   &      0.8\\
        \hline
        \end{tabular}
\label{tab:reg}
    \end{table*}

\smallskip
\noindent
{\em Model as Prior:}
The models can be used to induce an improper prior distribution
(or alternatively a penalty function) on the period $p$.
%Denote $\x_p$ as the modulo of $\x$ w.r.t.\ to period $p$, then
Given period $p$ and sample points $\x$
the prior is given by
      \begin{equation}
        \Pr(p) = \underset{i\in\{C,E,R\}}{\max}\left(\Pr(\y|\x,p; \mathcal{M}_i)\right)
      \end{equation}
where from the perspective of $\mathcal{M}_i$, $\x$ and corresponding points
in $\y$ are interpreted as if they were sampled modulo $p$.
%This is an improper prior because it is not an actual probability
%distribution.
Thus, combining this prior with the marginal likelihood, a
Maximum A Posteriori (MAP) estimation can be obtained. Adding a
regularization parameter $\gamma$
to obtain a tradeoff between the marginal likelihood and the improper prior
we get our criterion:
%[PP IS IT x_p modulo or not ? BW: what's x_p?
  \begin{equation}
  \label{eqn:regularize}
      \begin{split}
         \log\Pr(p|\x,\y;\mathcal{M}) &= \gamma\log\Pr(\y|\x, p;\mathcal{M}) \\
         &\quad + (1-\gamma)\log\Pr(p)
      \end{split}
\end{equation}
where $\Pr(\y|\x, p;\mathcal{M})$ is exactly as Equation~(\ref{eqn:malike})  where
the period portion of $\mathcal{M}$ is fixed to be $p$.
When using this approach with our algorithm we use
Equation~(\ref{eqn:regularize}) instead of Equation~(\ref{eqn:malike}) as
the score function in lines 5 and 13 of the algorithm.  The results for
different values of $\gamma$ (with subsampling and $5$ iterations) are shown
in Table~\ref{tab:reg}.  The results show that \textsc{gmt} on its own
($\gamma= 0$) is a good criterion for period finding.  This is as one might
expect because the OGLEII dataset includes only stars of the three types
captured by \textsc{gmt}.

In this experiment, regularized versions do not improve the result of the
\textsc{gmt} model. However, we believe that this will be the method of
choice in other cases when the prior information is less strong. In
particular, if the data includes unknown shapes that are not covered by the
generative model then the prior on its own will fail. On the other hand when
using Equation~(\ref{eqn:regularize}) with enough data the prior will be
dominated by the likelihood term and therefore the correct period can be
detected.
In contrast, the filter method
discussed next does not have such functionality.

\smallskip
\noindent
{\em Model as Filter:}
Our second approach
uses the model as a post-processing filter and
it is applicable to any method that scores different periods
before picking the top scoring one as its estimate.
For example, suppose we are
given the top $K$ best periods $\{p_i\}, i=1,\cdots, K$ found by LS, then
we choose the one such that
\begin{equation}
        p^* = \underset{i\in\{1,\cdots,K\}}{\text{argmax}}\left(\underset{j\in\{C,E,R\}}{\max}\left[\log\Pr(\y| \x, p_i; \mathcal{M}_j)\right]\right).
\end{equation}
Thus,
when using the \textsc{gmt} as a filter, step 17 in our algorithm is changed to
record the top $K$ frequencies from the last {\bf for} loop, evaluate each
one using the $\textsc{gmt}$ model likelihood, and output the top scoring
frequency.

\smallskip
\noindent
{\em Heuristic for Variable Periodic Stars:}
The two approaches above are general and can be used in any problem where a
model is available. For the astrophysics problem we develop another
heuristic that specifically addresses the half period problem of EBs. In
particular, when using the filter method, instead of choosing the top $K$
periods, we double the selected periods, evaluate both the original and
doubled periods $\{p_i, 2p_i\}$ using the \textsc{gmt} model, and
choose the best one.

Results of experiments using the filter method with and without the domain
specific heuristic are given in Table~\ref{tab:pen}, based on the 5 iteration
version of subsampling GP.
The filter method significantly improve the performance of our
algorithm showing its general applicability.
The
domain specific heuristic provides an additional improvement.
For LS,
the general filter method does not help but the
domain specific heuristic significantly improves its performance.
By analyzing the errors of both GP and LS, we found that
their error regions are different.
%they share only 60\% of the errors.
Therefore, we further propose a method that combines
the two methods in the following way: pick the top $K$ periods found by both
methods and evaluate the original and doubled periods using the \textsc{gmt} to
select the best one. As Table~\ref{tab:pen} shows, the combination gives an
additional 2\% improvement on the OGLEII subset.

      \begin{table}[t]
        \caption{Comparisons of different algorithms on OGLEII subset using
          the \textsc{GMT} as a filter. \textsc{Single} denotes without the double period heuristic.}
        \centering
        \begin{tabular}{||c||c|c|c||}
        \hline
        &  \textsc{original} & \textsc{single filter} & \textsc{filter} \\
        \hline
        \textsc{ls} &  0.7333  & 0.7243 & \textbf{0.9053} \\
        \hline
        \textsc{gp} & 0.8000 & 0.8829 & \textbf{0.9081} \\
        \hline
        \textsc{ls+gp} & - &0.8811 &\textbf{0.9297}\\
        \hline
        \end{tabular}
        \label{tab:pen}
        \end{table}

\begin{table*}[t]
\caption{Comparisons of accuracies for full set of OGLEII.}
\centering
\begin{tabular}{||c||c|c|c|c||}
        \hline
        &  \textsc{method in~\citep{wachman2009thesis}} & \textsc{ls-filter} & \textsc{gp-filter} & \textsc{gp-ls-filter} \\
        \hline
        \textsc{acc} &  0.8680  & $0.8975\pm 0.04$ & $0.8963\pm 0.03$ & $\mathbf{0.9243\pm 0.03}$ \\
        \hline
        \end{tabular}
\label{tab:cogleacc}
\end{table*}

\subsubsection{Application}
Finally, we apply our method using marginal likelihood with two level grid
search, sub-sampling at 15\%, 2 iterations, and filtering on the
complete OGLEII data set with 13974 instances minus the development OGLEII
subset. Note that the parameters of the algorithm, other than domain
dependent heuristics, are chosen based on our results from the artificial
data.  The accuracy is reported using 10-fold cross validation under the following setting:
the \textsc{gmt} is trained using the training set and we seek to find the
periods for the stars in the test set. We compare our results to the best
result from~\citep{wachman2009thesis} that used an improvement of LS, despite
the fact that they filtered out 1719 difficult stars due to insufficient
sampling points and noise.
% while we include all the dataset.
The results are
shown in Table~\ref{tab:cogleacc}. We can see that our approach significantly
outperforms existing methods on OGLEII.
%As shown by \cite{wachman2009thesis} correct identification of periods is
%important for automatic categorization of stars using machine learning.

%% conclusion part
\section{Related Work}
\label{sec:relatedwork}

Period detection has been extensively studied in the literature and
especially in astrophysics. The periodogram, as a tool for spectral
analysis, dates back to the 19th century when Schuster applied it to
the analysis of some data sets. The behavior of the periodogram
in estimating frequency was discussed by
\cite{deeming1975fourier}.
%whose construction which is often referred as the
%Deeming Periodogram in Astrophysics literature.
The periodogram is
defined as the modulus-squared of its discrete Fourier
transform~\citep{deeming1975fourier}. \cite{lomb1976least} and
\cite{scargle1982studies} introduced the so-called
Lomb-Scargle (LS) Periodogram that was discussed above and which
rates periods based on the sum-of-squares error of a sine wave at
the given period.  This method has been used in astrophysics
\citep{cumming2004detectability,wachman2009thesis} and has also been
used in
Bioinformatics~\citep{glynn2006detecting,wentao2008detecting}.
One can show that the LS periodogram is identical to the equation we would derive
if we attempted to estimate the harmonic content of a data set at a specific
frequency using the linear least-squares model~\citep{scargle1982studies}.
This technique was originally named least-squares spectral analysis
method \cite{vanicek1969approximate}.
Many extensions of the LS periodogram exist in the
literature~\citep{bretthorst2001generalizing}.  \cite{hall2006using}
proposed the periodogram for non-parametric regression models and discussed
its statistical properties. This was later applied to the situation where the
regression model is the superposition of functions with different
period~\citep{hall2008nonparametric}.

The other main approach uses least-squares estimates, equivalent to maximum
likelihood methods under Gaussian noise assumption, using different choices
of periodic regression models. This approach, using
finite-parameter trigonometric series of different orders,
has been explored by various authors
\citep{hartley1949tests,
 quinn1991estimating, quinn1991fast, quinn1999fast, quinn2001estimation}.
Notice that if the order of the
trigonometric series is high then
this is very close to
nonparametric methods~\citep{hall2008nonparametric}.

Another intuition is to minimize some measure of dispersion of the
data in phase space. Phase Dispersion Minimization
\citep{stellingwerf1978period}, described above, performs a least
squares fit to the mean curve defined by averaging points in bins.
%divides a cycle into (possibly overlapping) phase bins and calculates the
%$\chi^2$ agreement among those data points that fall into each bin at the
%trial period.
\cite{lafler1965rr} described a procedure which involves trial-period
folding followed by a minimization of the differences between observations of
adjacent phases.

Other least squares methods use smoothing based on splines, robust splines, or
variable-span smoothers.
%\citep{craven1978smoothing,oh2002period,mcdonald1986periodic}.
\cite{craven1978smoothing} discussed the problem of smoothing
periodic curve with spline functions in the regularization framework
and invented the generalized cross-Validation (GCV) score to
estimate the period of a variable star. \cite{oh2002period}
extended it by substituting the smoothing splines with robust
splines to alleviate the effects caused by outliers. Supersmoother,
a variable-span smoother based on running linear smooths, is
used for frequency estimation in~\citep{mcdonald1986periodic}.

Several other approaches exist in the literature.
Perhaps the most related work is \citep{hall2000nonparametric} who
studied nonparametric models for frequency estimation, including the
Nadaraya-Watson estimator, and discussed their statistical
properties. This was extended to perform inference for multi-period
functions~\citep{hall2003nonparametric} and evolving periodic
functions~\citep{genton2007statistical,hall2008nonparametric}. Our
work differs from~\citep{hall2000nonparametric} in three aspects: 1)
the GP framework presented in this paper is more general in that one
can plug in different periodic covariance functions for different prior
assumptions; 2) we use marginal likelihood that can be interpreted
to indicate how the data agrees with our prior belief; 3) we
introduce mechanisms to overcome the computational complexity of
period selection.

Other approaches include
%Analysis of Variance (ANOVA),
entropy minimization \citep{Huijse2011}, data
compensated discrete Fourier transform \citep{ferraz1981estimation},
and Bayesian models \citep{gregory1996bayesian,scargle1998studies}.
Recently, Bayesian methods have also been applied to solve the
frequency estimation problem, such as Bayesian binning for
Poisson-regime~\citep{gregory1996bayesian} and Bayesian
blocks~\citep{scargle1998studies}.

\section{Conclusion}

The paper introduces a nonparametric Bayesian approach for period estimation
based on Gaussian process regression. We develop a model selection algorithm
for GP regression that combines gradient based search and grid search, and
incorporates several algorithmic improvements and approximations leading to a
considerable decrease in run time. The algorithm performs significantly
better than existing state of the art algorithms when the data is not
sinusoidal.  Further, we show how domain knowledge can be incorporated into
our model as a prior or post-processing filter, and apply this idea in the
astrophysics domain.  Our algorithm delivers significantly higher accuracy
than existing state of the art in estimating the periods of variable periodic
stars.

%In this paper, we investigate the period finding problem and present a
%nonparametric Bayesian approach based on Gaussian process regression
%models. To deal with the unique problem of period estimation, we propose a
%model selection method for GP regression that combines gradient based search
%and grid search. We also propose several algorithmic improvements and
%approximations leading to a considerable decrease in run time. Further, we
%show how domain knowledge can be incorporated into our model. We perform
%extensive experiments on both synthetic data and astrophysics time series to
%demonstrate the performance of the proposed method.  Compared to previous
%work our methods provide significantly better accuracy on the astrophysics
%data.

An important direction for future work is to extend our model to develop a
corresponding statistical test for periodicity, that is, to determine
whether a time series is periodic.
This will streamline the application of our algorithm to new
astrophysics catalogs such as
MACHO~\citep{Alcock1993} where both periodicity testing and period estimation
are needed.
Another important direction is establishing the
theoretical properties of our method.  \cite{hall2000nonparametric}
provided the first-order properties of nonparametric estimators such that
under mild regularity conditions, the estimator is consistent and
asymptotically normally distributed. Our method differs in two ways: we use a
GP regressor instead of Nadaraya-Watson estimator, and we choose the period
that minimizes marginal likelihood rather than using a cross-validation
estimate. Based on the well known connection between kernel regression and GP
regression, we conjecture that similar results exist for the proposed method.

\section*{Acknowledgments}
This research was partly supported by NSF grant IIS-0803409. The
experiments in this paper were performed on the Odyssey cluster
supported by the FAS Research Computing Group at Harvard and the
Tufts Linux Research Cluster supported by Tufts UIT Research
Computing.

%This research was partly supported by NSF grant IIS-0803409.
%The experiments in this paper were performed on the
%Odyssey cluster at
%Harvard University and the Tufts Linux Research Cluster supported by Tufts UIT Research Computing.

%\clearpage
%\newpage
\appendix
\section{Low rank approximation}

In this appendix, we complete the details on how the first order approximation with low rank approximation can be
achieved by a series of rank one updates/downdates of the Cholesky
factors. As shown by~\cite{seeger2007low} each such update can be
done in $\fO(N^2)$ using a series of Givens rotations.

It can be easily seen that $\widetilde{\K}$ is a real symmetric
matrix. Denote its eigendecomposition as
$\widetilde{\K}=\boldsymbol{U}\boldsymbol{\Lambda}\boldsymbol{U}^T$,
then it can be written as the sum of a series of rank one
components,
 \begin{equation}
   \widetilde{\K} = \sum_{i=1}^N \text{sgn}(\lambda_i)
   \left(\sqrt{|\lambda_i|}\mathbf{u}_i\right)\left(\sqrt{|\lambda_i|}\mathbf{u}_i\right)^T
 \end{equation}
 where $\lambda_i$ is the $i$th eigenvalue and $\mathbf{u}_i$ is the corresponding eigenvector.
 Furthermore, we perform a low rank approximation to $\widetilde{\K}$ such that
 \begin{equation}
   \widetilde{\K} \approx \sum_{i=1}^M \text{sgn}(\lambda_{(i)})\left(\sqrt{|\lambda_{(i)}|}\mathbf{u}_{(i)}\right)\left(\sqrt{|\lambda_{(i)}|}\mathbf{u}_{(i)}\right)^T
 \end{equation}
where $M < N$ is a predefined rank and $\lambda_{(i)}$ and
$\mathbf{u}_{(i)}$ are the $i$th largest (in absolute value)
eigenvalue and its corresponding eigenvector. Therefore we have,
\begin{equation}
  \K_{w_1} \approx \bL\bL^T + \sum_{i=1}^M \text{sgn}(\lambda_{(i)})((\Delta w)^{1/2}\boldsymbol{\ell}_{i})((\Delta w)^{1/2}\boldsymbol{\ell}_{i})^T
\end{equation}
where $\boldsymbol{\ell}_i =
\sqrt{|\lambda_{(i)}|}\mathbf{u}_{(i)}$. We can see that the
complexity for calculating the Cholesky factor of $\K_{w_1}$ becomes
$\fO(MN^2)$. Therefore, we can choose an $\epsilon$-net
$\mathcal{E}$ of the fine grid such that $\forall w \in \mathcal{F},
\sup_{v\in\mathcal{E}}|w - v| < \epsilon$, perform the exact
Cholesky decomposition directly only on the $\epsilon$-net, and use
the approximation on the other frequencies.
%Thus instead of
%decomposing kernel matrices for all frequencies that belong to
%$\mathcal{F}$, we reduced such operations to be performed only
%$2\times |\mathcal{E}|$ times.
In this way we reduce the complexity from $\fO(|\mathcal{F}| N^3)$ to
$\fO(|\mathcal{E}| N^3 + |\mathcal{F}| MN^2 ) $.

\bibliography{pfind}{}

\end{document}